\documentclass[10pt,twocolumn,letterpaper]{article}

\usepackage[pagenumbers]{cvpr} % To force page numbers, e.g. for an arXiv version

\usepackage[dvipsnames]{xcolor}

\usepackage{multirow}
\usepackage{times}
\usepackage{epsfig}
\usepackage{graphicx}
\usepackage{amsmath}
\usepackage{amssymb}
\usepackage{stfloats}
\usepackage{float}
\usepackage[english]{babel}
\usepackage[font=small,labelfont=bf]{caption}
\usepackage{float}
\usepackage{graphicx}
\usepackage{amsmath}
\usepackage{color}
\usepackage{array,multirow}
\usepackage{colortbl}
\usepackage{hhline}
\usepackage{multirow}
\usepackage{amsmath}
\usepackage[dvipsnames]{xcolor}
\usepackage{bbm}
\usepackage{mathtools}
\usepackage{enumitem,kantlipsum}
\usepackage[normalem]{ulem}

\newcommand{\argmin}[1]{\underset{#1}{\operatorname{argmin}}\;}

\usepackage{dsfont}
\usepackage{enumitem}
\usepackage{algorithm2e}
\def\multiset#1#2{\ensuremath{\left(\kern-.3em\left(\genfrac{}{}{0pt}{}{#1}{#2}\right)\kern-.3em\right)}}
\usepackage{booktabs}
\usepackage{xcolor}
\usepackage[sort&compress, numbers]{natbib}

\SetCommentSty{mycommfont}
\usepackage{pifont}
\def\methodName{PiRO}
\def\archName{PAN}
\usepackage{balance}
\definecolor{cvprblue}{rgb}{0.21,0.49,0.74}
\usepackage[pagebackref,breaklinks,colorlinks,citecolor=cvprblue]{hyperref}

\title{Dual Pose-invariant Embeddings: Learning Category and Object-specific Discriminative Representations for Recognition and Retrieval}

\author{Rohan Sarkar, \hspace{0.05in} Avinash Kak\\
Electrical and Computer Engineering, Purdue University, USA\\
{\tt\small \{sarkarr, kak\}@purdue.edu}
}

\begin{document}
\maketitle
\begin{abstract}

In the context of pose-invariant object recognition and retrieval, we
demonstrate that it is possible to achieve significant improvements in
performance if both the category-based and the object-identity-based
embeddings are learned simultaneously during training.
In hindsight, that sounds intuitive because learning about the
categories is more fundamental than learning about the individual
objects that correspond to those categories. However, to the best of
what we know, no prior work in pose-invariant learning has
demonstrated this effect. This paper presents an attention-based
dual-encoder architecture with specially designed loss functions that
optimize the inter- and intra-class distances simultaneously in two
different embedding spaces, one for the category embeddings and the
other for the object level embeddings. The loss functions we have
proposed are pose-invariant ranking losses that are designed to
minimize the intra-class distances and maximize the inter-class
distances in the dual representation spaces.   We demonstrate the power
of our approach with three challenging multi-view datasets, ModelNet-40, ObjectPI, and FG3D. 
With our dual approach, for single-view object
recognition, we outperform the previous best by 20.0\% on ModelNet40, 2.0\% on ObjectPI, and 46.5\% on FG3D.  On the other
hand, for single-view object retrieval, we outperform the previous
best by 33.7\% on ModelNet40, 18.8\% on ObjectPI, and 56.9\% on FG3D.

\end{abstract}

\section{Introduction}

Pose-invariant recognition and retrieval \cite{PIE2019} is an
important problem in computer vision with practical applications in
robotic automation, automatic checkout systems, and inventory
management. The appearance of many objects belonging to the same
general category can vary significantly from different viewpoints,
and, yet, humans have no difficulty in recognizing them from arbitrary
viewpoints.  In pose-invariant recognition and retrieval, the focus is
on mapping the object images to embedding vectors such that the
embeddings for the objects that belong to the same category are pulled
together for all the available viewpoints in relation to the
embeddings for the objects for the different categories.

Our work demonstrates that the performance of
pose-invariant learning as described above can be significantly
improved if we disentangle the category-based learning from the
object-identity-based learning.

\begin{figure}
    \centering
    \includegraphics[width=0.4\textwidth]{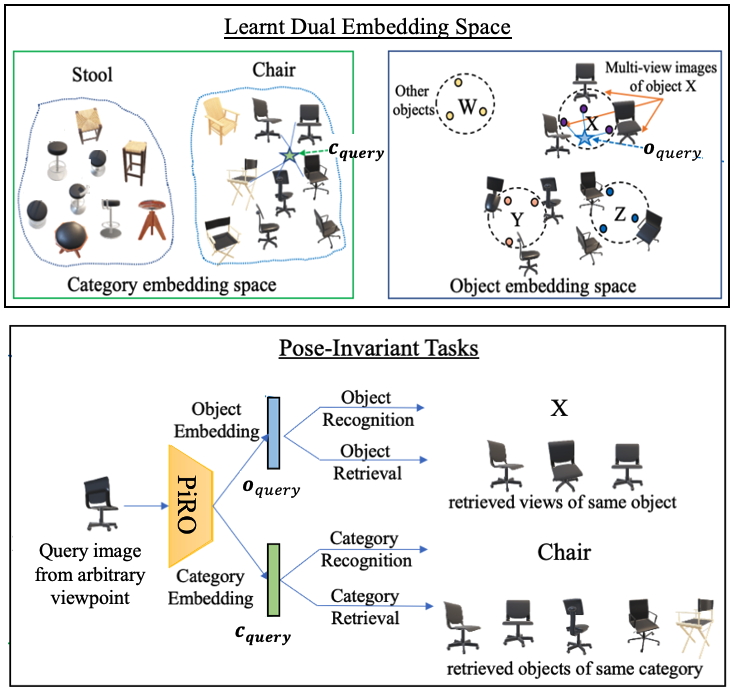}
    \caption{\em The upper panel shows objects belonging to two
      different categories, chair and stool. In the proposed
      disentangled dual-space learning, the goal for the learning of
      category-based embeddings is to capture what maximally
      discriminates the objects belonging to the two categories --- the
      presence or the absence of the back-rest.  On the other hand,
      the object-identity based embeddings are meant to capture what
      is distinctive about each object. 
      The lower panel illustrates our dual-space approach for simultaneously learning the embeddings in two different spaces for category and object-identity-based recognition and retrieval tasks. }
      \vspace{-0.1in}
    \label{fig:intr_diag}
    \vspace{-0.1in}
\end{figure}

Fig. \ref{fig:intr_diag} illustrates what we mean by disentangling the
category-based representation from the object-identity-based
representation.  Assume that an object database contains images of
different types of chairs and different types of stools. We would want
our network to learn the category-based embedding vectors for the
chair class and for the stool class.  These embeddings need to capture
what is maximally discriminating between the chairs and the stools ---
the presence or the absence of a back-rest.  At the same time, we
would want the network to learn object-identity based embeddings.
These embeddings should represent what is distinctive about each
object type in relation to all other objects types in the same
category.  For example, in the chair category, we would want the
network to be able to discriminate between, say, lounge chairs and
desk chairs.

Prior work \cite{mvcnn15, RotationNet2018, TCL2018, ModelNet40} has
employed multi-view deep networks to learn aggregated multi-view
representations capturing the variability in object appearance under
different pose transformations.  While these methods demonstrate good
performance in category-level tasks when multiple views of objects are
available during inference, {\em their performance degrades when only
  a single view is available}. Since real-world applications often
necessitate inference from single views, Ho et al. \cite{PIE2019}
proposed a family of pose-invariant embeddings for both recognition
and retrieval by imposing constraints such that the single-view
embeddings of an object are clustered around its multi-view
embeddings, which in turn are clustered around a proxy embedding
representing the associated high-level category that the object
belongs to.  However, this approach does not do a good enough job of
separating the embeddings for two different objects that belong to the
same category (e.g., two different types of chairs, two different
types of kettles, etc.), As a result, prior approaches perform well on
category-level tasks but not on object-level tasks, as we will
demonstrate later in our experimental results (see Tables
\ref{tbl:resultsPICR}, \ref{tbl:fg3d_comp}).

Here is arguably the most significant difference between the previous
methods and the one being proposed in this paper: {\em Rather than
  learning representations that capture both category-specific and
  object-specific discriminative features within the same embedding
  space, we simultaneously learn them in two distinct embedding
  spaces}, as depicted in the lower panel in
Fig. \ref{fig:intr_diag}. In one space that is devoted to
category-based representations, objects from the same category can be
closely embedded together, capturing shared characteristics among
them, while in the other space, the one for object identity-based
representations, embeddings for the different object types (within
the same category or otherwise) are allowed to be as separated as
dictated by the attributes that differentiate them. This strategy
enables our network to learn object representations that are more
discriminative overall. This should explain the superior performance
of our framework in both recognition and retrieval, especially for the
more difficult case when only a single-viewpoint query image is
available.  For single-view
object recognition, we get an improvement in accuracy of 20.0\% on
ModelNet40, 46.5\% on FG3D, and 2.0\% on ObjectPI. Along the same lines, for the case
of single-view object retrieval, we achieve a significant mAP
improvement of 33.7\% on ModelNet-40, 56.9\% on FG3D, and 18.8\% on ObjectPI datasets.

In order to learn the dual embeddings simultaneously, we propose an
encoder that we refer to as the Pose-invariant Attention Network
(PAN). PAN uses a shared CNN backbone for capturing visual features
common to both the category and the object-identity based
representations from a set of images of an object recorded from
different viewpoints. The visual features are then mapped to separate
low-dimensional category and object-identity based embeddings using
two fully connected layers. PAN also aggregates visual features of
objects from different views using self-attention to generate what we
call multi-view embeddings. The dual embeddings, defined in Section
\ref{sec:approach}, can be used for {\em both category and
  object-level recognition and retrieval from single and multiple
  views.}

For training the network, we propose two pose-invariant category and
object-identity based losses that are jointly optimized to learn the
dual embeddings.  The pose-invariant category loss clusters together
the instances of different objects belonging to the same category
while separating apart the instances from different categories in the
category embedding space.  On the other hand, the pose-invariant
object-identity based loss clusters together the instances that carry
the same object-identity label and separates what would otherwise be
mutually confusing object instances with two different object-identity
labels from the same category in the object embedding space.

\section{Background and Related work}
\label{sec:lit_review}
\noindent 
{\bf (A) Ranking and proxy-based losses:} Ranking losses, used in deep metric learning, focus on optimizing the relative pair-wise distances between exemplars (pairs \cite{Contrastive2005}, triplets \cite{Triplet2015} or quadruplets \cite{ReID2017}), such that similar samples are pulled closer and dissimilar samples are pushed apart.   
For ranking losses, the selection of informative exemplars \cite{HardUseful2020, LSADE2016, Schroff2015, FaceRec2015, LSFE2016, SmartMining2017, HTL2018, XBM2020, CBHEM2019} is crucial, which however incurs additional computational costs and memory. 
To reduce the training complexity, proxy-based approaches \cite{ProxyNCA} define a proxy embedding for each class and optimize sample-to-proxy distances. %as  
%$L_{prx}(\mathbf{x}, c;\mathbf{P}) =  -log \frac{e^{-d(g(\mathbf{x}), \mathbf{p_c})}}{\sum\limits_{k \neq c}e^{-d(g(\mathbf{x}), \mathbf{p_k})}}$, for an image embedding $g(\mathbf{x})$ from class $c$. 
However, they only capture relationships between samples and the proxies, which are less informative compared to the extensive sample-to-sample relations inherent in pair-based losses, which is particularly important for fine-grained tasks.

\noindent 
{\bf (B) Multi-view and Pose-Invariant Classification and Retrieval: }
In multi-view object recognition and retrieval \cite{mvcnn15, RotationNet2018, TCL2018, ModelNet40}, each object from category $c$ is captured from a set of $V$ views and is denoted by  $\mathbf{X} = \{\mathbf{x}_k\}_{k=1}^{V}$. For each object, a set of single-view embeddings are extracted by inputing each image $\mathbf{x}_k$ to a network $g_s$, which are then aggregated to generate multi-view embeddings as $g_m(\mathbf{X}) = \Phi(\{g_s(\mathbf{x}_k)\}_{k=1}^{V})$, where $\Phi$ denotes the aggregation operation. Multi-view losses cluster the multi-view embeddings of objects from the same category together and yield good performance on category-based tasks when multiple views of objects are available during inference. However, they perform poorly when only a single view is available during inference as the single-view embeddings are not constrained to be close to the multi-view embeddings in the embedding space. To mitigate this, the approach by \cite{PIE2019} learns pose-invariant embeddings by combining two separate view-to-object and object-to-category models trained using different types of pose-invariant losses. These losses optimize the pose-invariance distance defined as
\begin{equation}
d^{pi}(\mathbf{x}, \mathbf{X}, \mathbf{p}_c) = \alpha d(g_s(\mathbf{x}), g_m(\mathbf{X})) + \beta d(g_m(\mathbf{X}), \mathbf{p}_c)
\label{eqn:d_pi}
\end{equation}
where, $\alpha$ promotes the clustering of single-view embeddings around the object's multi-view embedding, while $\beta$ encourages the clustering of the multi-view embedding of the object around the learned proxy embedding $\mathbf{p}_c$ for its category $c$. 
However, these losses do not effectively separate embeddings of distinct objects from the same category, as we will demonstrate later in Fig. \ref{fig:viscomp}. This results in poor performance on object-based tasks.

In summary, prior work focused primarily on
 learning category-specific embeddings, with the object-to-object
 variations within each category represented by the variations in the embedding vectors within the same embedding space. 
In contrast, we learn a unified model that explicitly decouples the object and category embeddings. The model is trained jointly using two proposed pose-invariant ranking losses. In the category embedding space, the proposed loss clusters instances of different objects belonging to the same category together. In the object embedding space, the proposed loss clusters different views of the same object while separating confusing instances of different objects from the same category, thereby capturing discriminatory features to distinguish between similar objects from the same category. This significantly improves object recognition and retrieval performance over prior methods (ref. Tables \ref{tbl:resultsPICR}, \ref{tbl:fg3d_comp}).

\noindent
{\bf (C) Attention-based architectures:} 
Since the advent of ViT \cite{ViT2021}, transformers have become increasingly popular for a variety of computer vision tasks. Most relevant to our work are hybrid architectures comprising a CNN backbone in conjunction with a transformer encoder that use multi-head attention layers to learn aggregated representations from image collections comprising different items \cite{Sarkar_2023_WACV, Sarkar_2022_CVPR} and multi-view 3D shape representations \cite{DAN2021, CVR2021} for classification and retrieval tasks. In contrast, we only use a single-head self-attention layer for each subspace to aggregate visual features extracted from a DNN across different views to learn multi-view embeddings. 
The architectures in \cite{DAN2021, CVR2021} learn multi-view shape representations for category-based tasks and require multi-view images at inference time. In contrast, our dual-encoder is designed to simultaneously learn pose-invariant category and object representations that can be utilized for both category and object-based tasks from single and multiple views during inference. 
Separate models for classification and retrieval tasks were proposed in \cite{Sarkar_2023_WACV}, whereas our unified model can address both tasks jointly. Also, positional encodings are utilized by \cite{ViT2021, DAN2021} to preserve input order, but not by \cite{ Sarkar_2023_WACV}. We omit positional encodings to ensure that the learned representations are independent of the input view order.

\section{Proposed Approach}
\label{sec:approach}

\begin{figure}[b]
      \centering
          \includegraphics[width=0.475\textwidth]{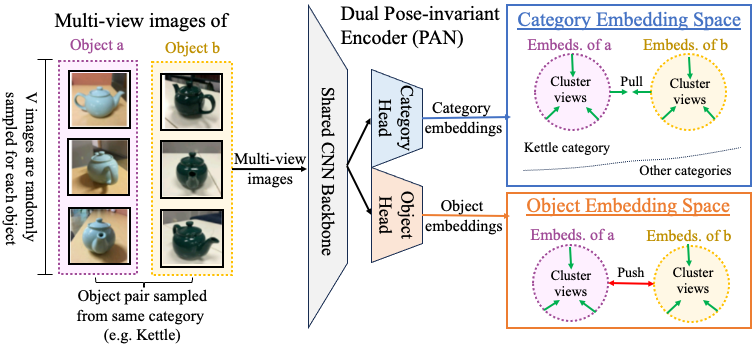}
          \vspace{-0.22in}
          \caption{\em An overview of our \methodName{} framework to learn the dual pose-invariant object and category embeddings using losses specifically designed for each embedding space. Multi-view images of two randomly chosen objects from the same category are used to learn common characteristics of the objects in the category embedding space and discriminatory attributes to distinguish between them in the object embedding space.} 
          \label{fig:learn}
          \vspace{-0.1in}
  \end{figure}

A high-level overview of our framework \methodName{} for learning dual
pose-invariant representations of objects is shown in
Fig. \ref{fig:learn}.  Our approach learns by comparing pairs of
objects belonging to the same category, while taking into account
their multi-view appearances.  This is illustrated in
Fig. \ref{fig:learn} where we show two different kettles, obviously
belonging to the same category, and, in the depiction in the figure,
we use three randomly chosen viewpoint images for each kettle.  For
the purpose of explanation, we have labeled the two objects as $a$
and $b$.  In general, we choose $V$ number of randomly selected
images from the different viewpoints for each object.
The objective of this within-category learning is to become aware of the
common attributes shared by these objects, like the spout, lid,
handle, and overall body structure, enabling their categorization as a
kettle.

The multi-view images are input to our proposed dual-encoder PAN,
which we introduce in Section \ref{sec:mvencoder}. The dual encoder
consists of a shared CNN backbone responsible for capturing common
visual features, along with two distinct heads dedicated to the
learning of the dual category and object-identity based embeddings.

The encoder is trained jointly using pose-invariant losses designed
for each respective embedding space, as described in Section
\ref{sec:PIOBJloss}. In the category embedding space, the loss is
designed to cluster together the embeddings of the objects from the
same category regardless of the viewpoints,
as shown in top-right of Fig. \ref{fig:learn}.  On the other hand, in the
object-identity embedding space, the loss is designed to cluster
together the embeddings for the instances that carry the same
object-identity label, again regardless of the viewpoints, while separating instances with different object-identity labels from the same category,
as shown in bottom-right of Fig. \ref{fig:learn}.  
 The idea is for the
encoder to capture shared characteristics among objects within the
same category in the category space and discriminatory attributes to distinguish between them in the object space.
These dual
embeddings can then be utilized for pose-invariant category and
object-based recognition and retrieval.

\noindent
\subsection{Pose-invariant Attention Network (\archName{}):}
\label{sec:mvencoder}
\begin{figure}[t]
		\centering
	\includegraphics[width=0.45\textwidth]{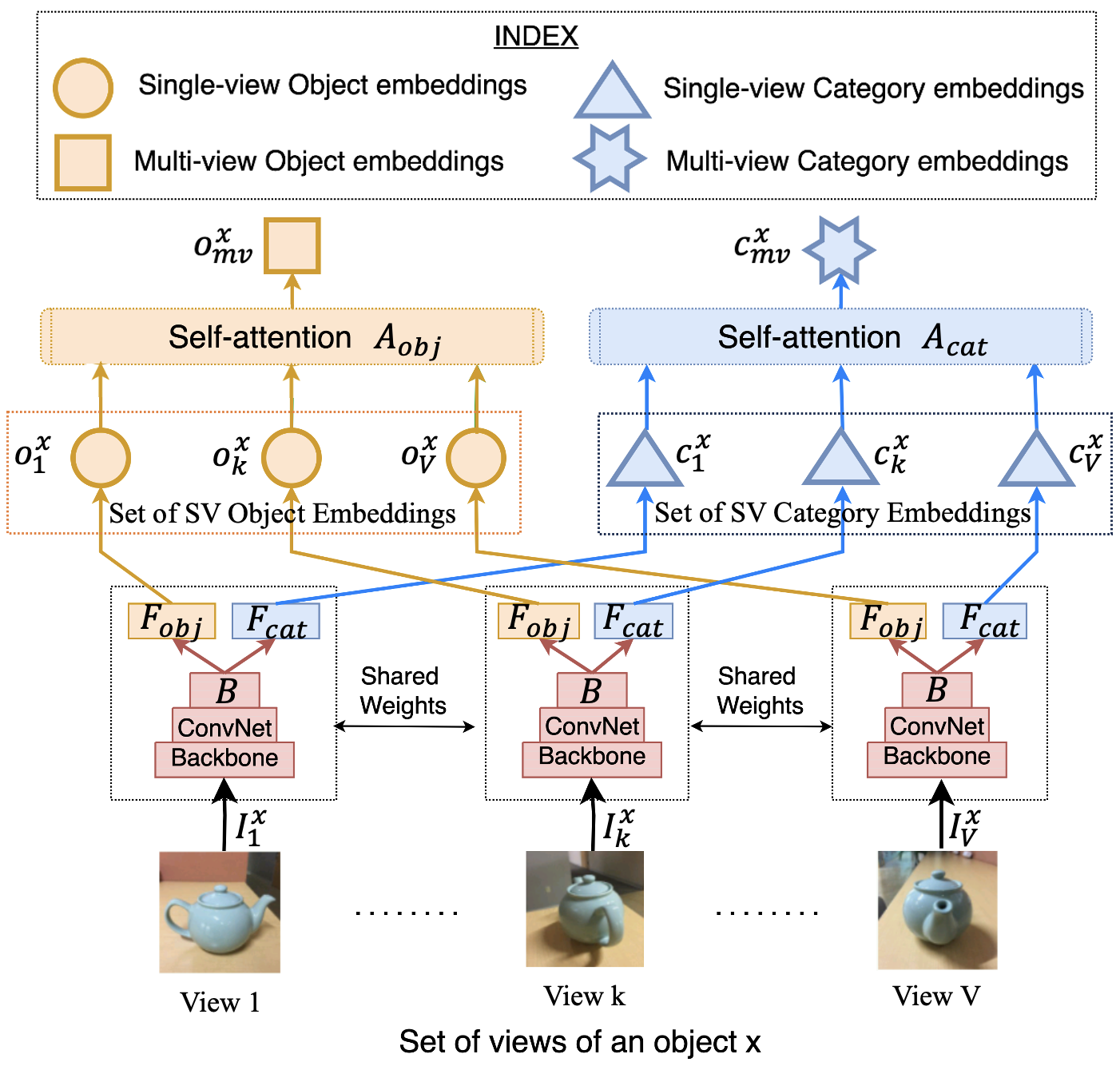}
 \vspace{-0.1in}
	\caption{\em The Pose-invariant Attention Network
          (\archName{}) takes a set of multi-view images of an object
          as input, producing both single-view and multi-view
          embeddings for each representational subspace. The object
          embeddings are depicted in orange, while the category
          embeddings are in blue.  }
	\label{fig:network}
 \vspace{-0.17in}
\end{figure}
Fig. \ref{fig:network} illustrates in greater detail the design of
PAN, the Pose-invariant Encoder shown previously in
Fig. \ref{fig:learn}.
It consists of a CNN backbone ($B$), two FC layers ($\mathit{F}_{obj}$
and $\mathit{F}_{cat}$) and two single-head self-attention layers
($\mathit{A}_{obj}$ and $\mathit{A}_{cat}$).  It takes as input an
unordered set of images from $V$ different views of an object $x$ from
category $l_x$ represented as $\mathbf{I}^x_{set} = \{\mathbf{I}^x_1,
\cdots, \mathbf{I}^x_k, \cdots, \mathbf{I}^x_V\}$.  The backbone and
FC layers for each view share the same weights.

The backbone learns visual features common to both the category and
object-identity representations.  The visual features extracted from each
object view are subsequently input to the FC layer
($\mathit{F}_{obj}$) to generate
%$d$-dimensional 
the object-identity embeddings. The set of {\em single-view object
  embeddings} for object $x$ is denoted by:
\begin{equation}
\mathcal{E}^x_{obj} = \{\mathbf{o}^x_k \mid \mathbf{o}^x_k = \mathit{F}_{obj}(B(\mathbf{I}^x_{k})) \hspace{0.1 in} \forall \mathbf{I}^x_k \in \mathbf{I}^x_{set} \} 
\label{eqn:objemb}
\end{equation}
Similarly, the shared visual features are input to another FC layer ($\mathit{F}_{cat}$) to generate %$d$-dimensional 
category embeddings. The set of {\em single-view category embeddings} for object $x$ is denoted by:
\begin{equation}
\mathcal{E}^x_{cat} = \{\mathbf{c}^x_k \mid \mathbf{c}^x_k = \mathit{F}_{cat}(B(\mathbf{I}^x_{k})) \hspace{0.1 in} \forall \mathbf{I}^x_k \in \mathbf{I}^x_{set} \} 
\label{eqn:catemb}
\end{equation}

The single-view object and category embeddings are then passed into
the self-attention layers $A_{obj}$ and $A_{cat}$ to learn the
corresponding multi-view embeddings. The self-attention mechanism
allows weighted interactions between the features extracted from one
view with features extracted from all the remaining views in the set
to capture the correlation between visual features across multiple
views effectively. The resulting feature vectors from the images of an
object are then aggregated using mean-pooling to get the multi-view
embeddings.
 The resulting {\em multi-view object and category embedding} for object $x$ is denoted by:
\begin{equation}
\mathbf{o}^{x}_{mv} = \frac{1}{V}\sum\limits_{k=1}^V \mathit{A}_{obj}(\mathcal{E}^x_{obj}),
\mathbf{c}^{x}_{mv} = \frac{1}{V}\sum\limits_{k=1}^V \mathit{A}_{cat}(\mathcal{E}^x_{cat})
\label{eqn:mvemb}
\end{equation}

\subsection{Pose-invariant Losses}
\label{sec:loss}
The single-view and multi-view embeddings extracted using PAN are used
for constructing pose-invariant losses that train the encoder to map
object images across different viewpoints to compact low-dimensional
subspaces, where the Euclidean distance between embeddings corresponds
to a measure of object similarity across viewpoints. We propose two
such pose-invariant losses for the object and category embedding
spaces next.

\begin{figure}[t]
		\centering
		\includegraphics[width=0.4\textwidth]{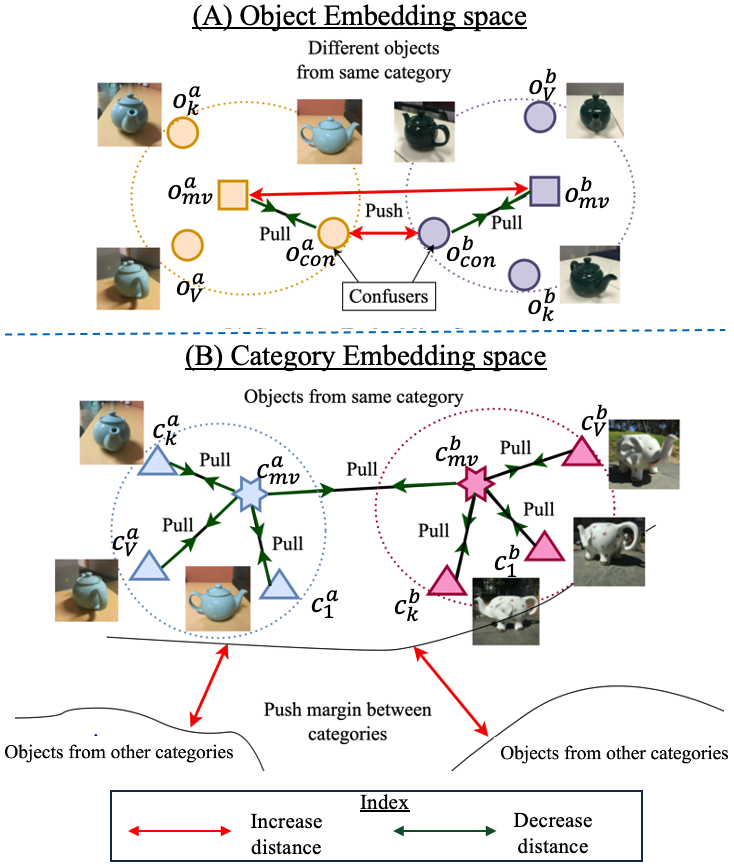}
  \vspace{-0.05in}
	\caption{\em The pose-invariant losses enhance intra-class compactness and inter-class separation in the dual embedding spaces. In the object embedding space (top), confusing instances of two different objects from the same category are separated. In the category embedding space (bottom), objects belonging to the same category are pulled closer while being separated from those belonging to other categories.}
	\label{fig:loss}
 \vspace{-0.1in}
\end{figure}

\noindent
{\em \bf (A) Pose-invariant Object Loss: }
\label{sec:PIOBJloss}
This loss is designed specifically for fine-grained object
recognition and retrieval from arbitrary viewpoints.  The loss
pulls together the embeddings of the different views of the same object,
as shown by the green arrows in Fig. \ref{fig:loss}(A). This allows
the encoder to learn common view-invariant features from multiple
views.  At the same time, it is designed to increase the inter-class
distances between the embeddings (as shown by the red arrows in the
same figure).  That allows the encoder to learn the discriminative
features to distinguish between visually similar objects from the same
category.

Let us consider a pair of objects $(a, b)$ from the same high-level
category as shown in Fig. \ref{fig:loss}(A).  The object-identity
embeddings generated by the encoder (ref. Eqn. \ref{eqn:objemb}) from
$V$ views for each of the objects are symbolically represented as the
two sets $\mathcal{E}^a_{obj}$ and $\mathcal{E}^b_{obj}$ respectively.
For each such pair, embeddings of different objects with the minimum
separation between them are the most informative and are chosen as the
{\em confusers}. These embeddings are called confusers because they
maximally violate the inter-class margin between the object pair and
are the most likely to confuse a classifier. The confusers denoted by
$\mathbf{o}^a_{con}$ and $\mathbf{o}^b_{con}$ are computed as
\begin{equation}
\mathbf{o}^{a}_{con}, \mathbf{o}^{b}_{con} = \argmin{\forall \mathbf{x} \in \mathcal{E}^a_{obj}, \forall \mathbf{y} \in \mathcal{E}^b_{obj}} d(\mathbf{x}, \mathbf{y})
\label{eqn:confuser}
\end{equation}
where, $d(\mathbf{x}, \mathbf{y}) = \|\mathbf{x} - \mathbf{y}\|_2$ is
the euclidean distance between the embeddings $\mathbf{x}$ and
$\mathbf{y}$.  The multi-view object embeddings $\mathbf{o}^{a}_{mv},
\mathbf{o}^{b}_{mv}$ from the respective object-identity classes are considered as
       {\em positives}.

The intra-class compactness and inter-class separability are controlled using two margins 
$\alpha$ and 
$\beta$ respectively.
Our pose-invariant object-identity loss has two components:\\
\noindent 
{\bf (i) Clustering loss} ensures that the distance between the
multi-view embedding and the single-view confuser embedding in
Eqn. \ref{eqn:confuser} of the same object-identity class $a$ does not
exceed the margin $\alpha$. For the object-identity class $a$, it is
defined as:
\begin{equation}
\mathcal{L}^a_{intra} = \Big[d(\mathbf{o}^{a}_{mv}, \mathbf{o}^{a}_{con})  - \alpha\Big]_+ 
\end{equation}
where, $[z]_+= max(z, 0)$ is the hinge loss.

\noindent 
{\bf (ii) Separation loss} ensures that the minimum distance between
the single-view confuser embeddings of two objects $a$ and $b$ and
also the separation between the multi-view object embeddings of the
corresponding objects is greater than a margin $\beta$. By separating
the confusers and multi-view embeddings of two objects from the same
category, the encoder will learn discriminatory features. It is
defined as:
\begin{equation}
\mathcal{L}^{a,b}_{inter} = \Big[ \beta - d(\mathbf{o}^a_{con}, \mathbf{o}^b_{con}) \Big]_+ + \Big[ \beta - d(\mathbf{o}^a_{mv}, \mathbf{o}^b_{mv}) \Big]_+
\end{equation}
The overall loss is defined as: 
\begin{equation}
\mathcal{L}^{a, b}_{piobj} = \mathcal{L}^a_{intra}+\mathcal{L}^b_{intra} +\mathcal{L}^{a, b}_{inter}
\label{eqn:piobj}
\end{equation} 
{\em \bf (B) Pose-invariant Category Loss: }
\label{sec:PICATloss}
As shown by the green arrows in Fig. \ref{fig:loss}(B), this loss
ensures that in the category embedding space, the single-view and
multi-view embeddings of an object are well clustered and the
multi-view embeddings for two different object-identity classes from
the same category are embedded close to each other and do not exceed a
margin $\theta$.  The clustering loss for the category embeddings (ref. Eqn. \ref{eqn:catemb}) of
the objects $a, b$ from the same category is defined as:
\begin{equation}
\mathcal{L}^{a,b}_{picat} = \Big[d^a_{sm} - \theta \Big]_+ + \Big[d^b_{sm} - \theta \Big]_+ + \Big[ d(\mathbf{c}^{a}_{mv}, \mathbf{c}^{b}_{mv}) -\theta \Big]_+ 
\end{equation}
where, $d^x_{sm} =
\frac{1}{V}\sum\limits_{k=1}^V\!d(\mathbf{c}^{x}_{k},
\mathbf{c}^{x}_{mv})$ is the mean of the distances between the
multi-view and single-view embeddings for an object $x$ in the
category embedding space.

\noindent
{\em \bf (C) Total loss: } In the category embedding space, we use the
large-margin softmax (L-Softmax) loss for separating the embeddings of
objects from different categories (shown by the red arrows in
Fig. \ref{fig:loss}(B)) using a margin $\gamma$. The dual-encoder PAN is
jointly trained using all the losses and the overall loss is defined
as
\begin{equation}
\mathcal{L} = \frac{1}{|\mathcal{P}|}\sum_{(a,b) \in \mathcal{P}} \mathcal{L}^a_{cat}+\mathcal{L}^b_{cat} + \mathcal{L}^{a, b}_{picat} + \mathcal{L}^{a, b}_{piobj} 
\end{equation}
where, $\mathcal{L}^x_{cat} = \frac{1}{V}\sum\limits_{k=1}^V
\mathcal{L}_{\gamma}(\mathbf{c}^{x}_k, l_x)$ such that
${L}_{\gamma}(\mathbf{c}^{x}_k, l_x)$ is the L-Softmax loss
\cite{LMSoftmax} with margin $\gamma$ for a category embedding
$\mathbf{c}^{x}_k$ of an object $x$ belonging to category $l_x$ from
any viewpoint $k$, and $\mathcal{P}$ is the set of all object pairs
where each pair is randomly sampled from the same category.

\section{Experiments}
\label{sec:results}
In this section, we evaluate our approach on pose-invariant classification and retrieval (PICR) tasks on three multi-view object datasets, report ablation studies at the end of this section, and additional results in the supplementary material. 

\noindent
{\bf (A) Setup, implementation details and results: }
\label{sec:exp_picr}

\begin{table*}[t]
\footnotesize 

\centering 
\setlength{\tabcolsep}{2pt}
\begin{tabular}{@{}lcccccccccccc@{}}
\toprule
\multirow{4}{*}{Dataset} & \multirow{4}{*}{\begin{tabular}[c]{@{}c@{}}Embed.\\ Space\end{tabular}} & \multicolumn{5}{c}{Classification (Accuracy \%)} & \multicolumn{5}{c}{Retrieval (mAP \%)} \\ \cmidrule(l){3-7} \cmidrule(l){8-12}
 &  & \multicolumn{2}{c}{Category} & \multicolumn{2}{c}{Object} & \multicolumn{1}{c}{\multirow{2}{*}{\bf Average}} & \multicolumn{2}{c}{Category} & \multicolumn{2}{c}{Object} & \multicolumn{1}{c}{\multirow{2}{*}{\bf Average }} \\ \cmidrule(lr){3-4} \cmidrule(lr){5-6} \cmidrule(lr){8-9} \cmidrule(lr){10-11}
 &  & \multicolumn{1}{c}{Single-view} & \multicolumn{1}{c}{Multi-view} & \multicolumn{1}{c}{Single-view} & \multicolumn{1}{c}{Multi-view} & \multicolumn{1}{c}{} & \multicolumn{1}{c}{Single-view} & \multicolumn{1}{c}{Multi-view} & \multicolumn{1}{c}{Single-view} & \multicolumn{1}{c}{Multi-view} & \multicolumn{1}{c}{} \\ \midrule 
\multirow{3}{*}{ObjectPI} & Single  & 69.56 $\pm$ 0.9 & 80.27 $\pm$ 1.9 & 88.35 $\pm$ 0.3 & 98.98 $\pm$ 0.9 & 84.29 $\pm$ 0.7 & 65.81 $\pm$ 0.5 & 75.60 $\pm$ 0.7 & 68.55 $\pm$ 0.5 & 99.46 $\pm$ 0.5 & 77.35 $\pm$ 0.4 \\
\cmidrule(l){2-12} 
 & Dual  & 70.22 $\pm$ 0.7 & 82.48 $\pm$ 1.0 & 93.07 $\pm$ 0.8 & 98.64 $\pm$ 0.5 & {\bf 86.10} $\pm$ {\bf 0.3} & 65.20 $\pm$ 0.4 & 82.80 $\pm$ 0.5
 & 80.61 $\pm$ 0.5 & 99.46 $\pm$ 0.3 & {\bf 82.02} $\pm$ {\bf 0.3} \\
 \midrule
\multirow{3}{*}{ModelNet} & Single  & 85.09 $\pm$ 0.3 & 88.08 $\pm$ 0.6 & 82.90 $\pm$ 1.5 & 86.75 $\pm$ 1.2 & 85.71 $\pm$ 0.5 & 78.88 $\pm$ 0.2 & 82.88 $\pm$ 0.2 
 & 61.89 $\pm$ 2.3 & 91.22 $\pm$ 0.8 & 78.71 $\pm$ 0.7 \\
\cmidrule(l){2-12} 
 & Dual  & 84.96 $\pm$ 0.2 & 88.32 $\pm$ 0.4 & 94.14 $\pm$ 0.3 & 96.88 $\pm$ 0.2 & {\bf 91.07} $\pm$ {\bf 0.2} & 79.30 $\pm$ 0.2 & 85.28 $\pm$ 0.4 & 84.46 $\pm$ 0.2 & 98.17 $\pm$ 0.1 & {\bf 86.80} $\pm$ {\bf 0.1} \\ \midrule 
\multirow{3}{*}{FG3D} & Single  & 78.18 $\pm$ 0.2 & 80.42 $\pm$ 0.1 & 26.51 $\pm$ 0.3 & 29.76 $\pm$ 0.7 & 53.72 $\pm$ 0.3 & 65.05 $\pm$ 0.3 & 69.28 $\pm$ 0.2 & 15.79 $\pm$ 0.1 & 41.98 $\pm$ 0.6 & 48.02 $\pm$ 0.3 \\
\cmidrule(l){2-12} 
 & Dual  & 78.89 $\pm$ 0.2 & 81.81 $\pm$ 0.1 & 83.00 $\pm$ 0.2 & 91.56 $\pm$ 0.1 & {\bf 83.81} $\pm$ {\bf 0.1} & 67.95 $\pm$ 0.3 & 74.24 $\pm$ 0.3 & 72.78 $\pm$ 0.3 & 95.47 $\pm$ 0.1 & {\bf 77.61} $\pm$ {\bf 0.2} \\
 \bottomrule
\end{tabular}
\vspace{-0.05in}
\caption{\em Pose-invariant Classification and Retrieval results on category and object-level tasks using our method for single and dual embedding spaces on the ModelNet-40, FG3D and ObjectPI datasets. The average performance along with standard deviation are reported. }
\label{tbl:single_dual}
\vspace{-0.15in}
\end{table*}

\noindent
{\bf Datasets: }
ModelNet-40 \cite{ModelNet40} is a multi-view dataset comprising  3983 objects (3183 train and 800 test) with roughly 100 unique CAD models per category from 40 common object categories. 
The dataset is generated by starting from an arbitrary pose and rotating each CAD model every 30 degrees resulting in 12 poses per model.  The ObjectPI dataset \cite{PIE2019} consists of images collected in the wild, by placing each object in a natural scene and capturing pictures from 8 views around the object, for 480 objects (382 train and 98 test) from 25 categories. We use the same training and test splits provided by \cite{PIE2019} for both datasets. Additionally, we also evaluate our method on FG3D \cite{FG3D} which is a large-scale dataset for fine-grained object recognition with 12 views per object for 25552 objects (21575 training and 3977 test) from 66 categories.\\ 
\noindent
{\bf Tasks: } Ho et al. \cite{PIE2019} proposed five tasks: 
{\em Single-view and multi-view category recognition}. These tasks predict the category from a single
view and a set of object views respectively. 
{\em Single-view and multi-view category retrieval}. The goal of these tasks is to retrieve images from the
same category as the query object from a single view and multiple views respectively.
{\em Single-view object retrieval}. This task aims to retrieve other views of the same object in the query view. 
We additionally report results using our method in Table \ref{tbl:single_dual} on three more tasks which are extensions of the above-mentioned tasks. These tasks are {\em single and multi-view object recognition} and {\em multi-view object retrieval}. The splits for evaluation on all these tasks will be publicly released. 
Classification and retrieval performance are reported as accuracy and mean average precision (mAP) respectively. \\
\noindent
{\bf Training details: }
 Images are resized to 224$\times$224 and normalized before being input to the network. 
 The VGG-16 network \cite{VGG} is used as the CNN backbone
 for a fair comparison with other state-of-the-art approaches. The last FC layers are modified to generate 2048-D embeddings and are initialized with random weights. A single layer and single head self-attention layer is used with a dropout of 0.25. The network is jointly trained using the proposed pose-invariant category and object losses. For all datasets, we set the margins $\alpha=0.25, \beta=1.00$ for the object embedding space and margins $\theta=0.25, \gamma=4.00$ for the category embedding space. 
 We use the Adam optimizer with a learning rate of $1e^{-5}$ for ObjectPI, ModelNet-40, and $5e^{-5}$ for FG3D. We train for 25 epochs and use the step scheduler that reduces the learning rate by half after every 5 epochs. \\
\noindent 
{\bf (B) Comparison with state-of-the-art: }
 \useunder{\uline}{\ul}{}
\begin{table*}[t]
\footnotesize
\centering 
\begin{tabular}{@{}l|llllllll|llllllll|@{}}
\toprule
\multirow{3}{*}{Method} & \multicolumn{8}{c|}{ModelNet-40 (12 views)} & \multicolumn{8}{c|}{ObjectPI (8 views)} \\ \cmidrule(l){2-9} \cmidrule(l){10-17}
 & \multicolumn{4}{l|}{Classification (Accuracy \%)} & \multicolumn{4}{c|}{Retrieval (mAP \%)} & \multicolumn{4}{l|}{Classification (Accuracy \%)} & \multicolumn{4}{c|}{Retrieval (mAP \%)} \\ \cmidrule(l){2-5}\cmidrule(l){6-9} \cmidrule(l){10-13}\cmidrule(l){14-17}
 & \multicolumn{1}{c}{\begin{tabular}[c]{@{}c@{}}SV\\ Cat\end{tabular}} & \multicolumn{1}{c}{\begin{tabular}[c]{@{}c@{}}MV\\ Cat\end{tabular}} & \multicolumn{1}{c}{\begin{tabular}[c]{@{}c@{}}SV\\ Obj\end{tabular}} & \multicolumn{1}{c|}{\bf Avg} & \multicolumn{1}{c}{\begin{tabular}[c]{@{}c@{}}SV\\ Cat\end{tabular}} & \multicolumn{1}{c}{\begin{tabular}[c]{@{}c@{}}MV\\ Cat\end{tabular}} & \multicolumn{1}{c}{\begin{tabular}[c]{@{}c@{}}SV\\ Obj\end{tabular}} & \multicolumn{1}{c|}{\bf Avg} & \multicolumn{1}{c}{\begin{tabular}[c]{@{}c@{}}SV\\ Cat\end{tabular}} & \multicolumn{1}{c}{\begin{tabular}[c]{@{}c@{}}MV\\ Cat\end{tabular}} & \multicolumn{1}{c}{\begin{tabular}[c]{@{}c@{}}SV\\ Obj\end{tabular}} & \multicolumn{1}{c|}{\bf Avg} & \multicolumn{1}{c}{\begin{tabular}[c]{@{}c@{}}SV\\ Cat\end{tabular}} & \multicolumn{1}{c}{\begin{tabular}[c]{@{}c@{}}MV\\ Cat\end{tabular}} & \multicolumn{1}{c}{\begin{tabular}[c]{@{}c@{}}SV\\ Obj\end{tabular}} & \multicolumn{1}{c|}{\bf Avg} \\ \midrule
MV-CNN & 71.0 & 87.9 & 65.6 & \multicolumn{1}{l|}{74.8} & 41.7 & 71.5 & 29.6 & 47.6 & 62.1 & 74.1 & 75.8 & \multicolumn{1}{l|}{70.7} & 53.8 & 72.3 & 42.6 & 56.2 \\
PI-CNN & \textbf{85.4} & 88.0 & 65.1 & \multicolumn{1}{l|}{79.5} & 77.5 & 81.8 & \textit{50.8} & \textit{70.0} & 66.5 & 76.5 & 61.6 & \multicolumn{1}{l|}{68.2} & 58.9 & 72.1 & 60.7 & 63.9 \\
MV-TC & 77.3 & {\ul 88.9} & 54.2 & \multicolumn{1}{l|}{73.5} & 63.5 & 84.0 & 36.6 & 61.4 & 65.7 & 79.2 & 65.9 & \multicolumn{1}{l|}{70.3} & 59.5 & {\it 77.3} & 51.8 & 62.9 \\
PI-TC & 81.2 & {\ul 88.9} & \textit{74.1} & \multicolumn{1}{l|}{\textit{81.4}} & 71.5 & {\it 84.2} & 41.4 & 65.7 & {\it 69.3} & 77.5 & {\ul 91.1} & \multicolumn{1}{l|}{{\it 79.3}} & {\it 63.8} & 76.7 & \textit{61.8} & {\it 67.4} \\
MV-Proxy & 79.7 & \textbf{89.6} & 37.1 & \multicolumn{1}{l|}{68.8} & 66.1 & {\ul 85.1} & 35.0 & 62.1 & 63.2 & 78.3 & 53.6 & \multicolumn{1}{l|}{65.0} & 57.9 & 74.7 & 49.3 & 60.6 \\
PI-Proxy & {\ul 85.1} & {\it 88.7} & 66.1 & \multicolumn{1}{l|}{80.0} & \textbf{79.9} & {\ul 85.1} & 40.6 & 68.5 & 68.7 & \textit{80.0} & 70.8 & \multicolumn{1}{l|}{73.2} & 62.6 & {\ul 78.2} & 49.4 & 63.4 \\ \midrule
\methodName-SE (Ours) & {\ul 85.1} & 88.1 & {\ul 82.9} & \multicolumn{1}{l|}{{\ul 85.4}} & \textit{78.9} & 82.9 & {\ul 61.9} & {\ul 74.6} & {\ul 69.6} & {\ul 80.3} & \textit{88.4} & \multicolumn{1}{l|}{{\ul 79.4}} & \textbf{65.8} & 75.6 & {\ul 68.5} & {\ul 70.0} \\
\methodName-DE (Ours) & {\it 85.0} &  88.3 & \textbf{94.1} & \multicolumn{1}{l|}{\textbf{89.1}} & {\ul 79.3} &  {\bf 85.3} & \textbf{84.5} & \textbf{83.0} & \textbf{70.2} & \textbf{82.5} & \textbf{93.1} & \multicolumn{1}{l|}{\textbf{81.9}} & {\ul 65.2} & {\bf 82.8} & \textbf{80.6} & \textbf{76.2} \\
\bottomrule
\end{tabular}
\vspace{-0.05in}
\caption{\em Comparison of performance on pose-invariant classification and retrieval tasks on the ObjectPI and ModelNet-40 datasets with the state-of-the-art approaches. The best, second-best, and third-best performance is highlighted in bold, underline, and italics respectively. The methods starting with MV indicate multi-view methods and those starting with PI indicate methods that learn pose invariant embeddings. For our method \methodName, SE and DE stands for single and dual embedding spaces. The average classification and retrieval performance indicate that we learn better representations for recognition and retrieval tasks on both datasets. The improvements in single-view object recognition and retrieval performance are the most significant. }
\label{tbl:resultsPICR}
\vspace{-0.15in}
\end{table*}
\begin{figure*}[b]
    \begin{minipage}{\textwidth}
		\centering
		\includegraphics[height=0.2\textwidth, width=0.245\textwidth]{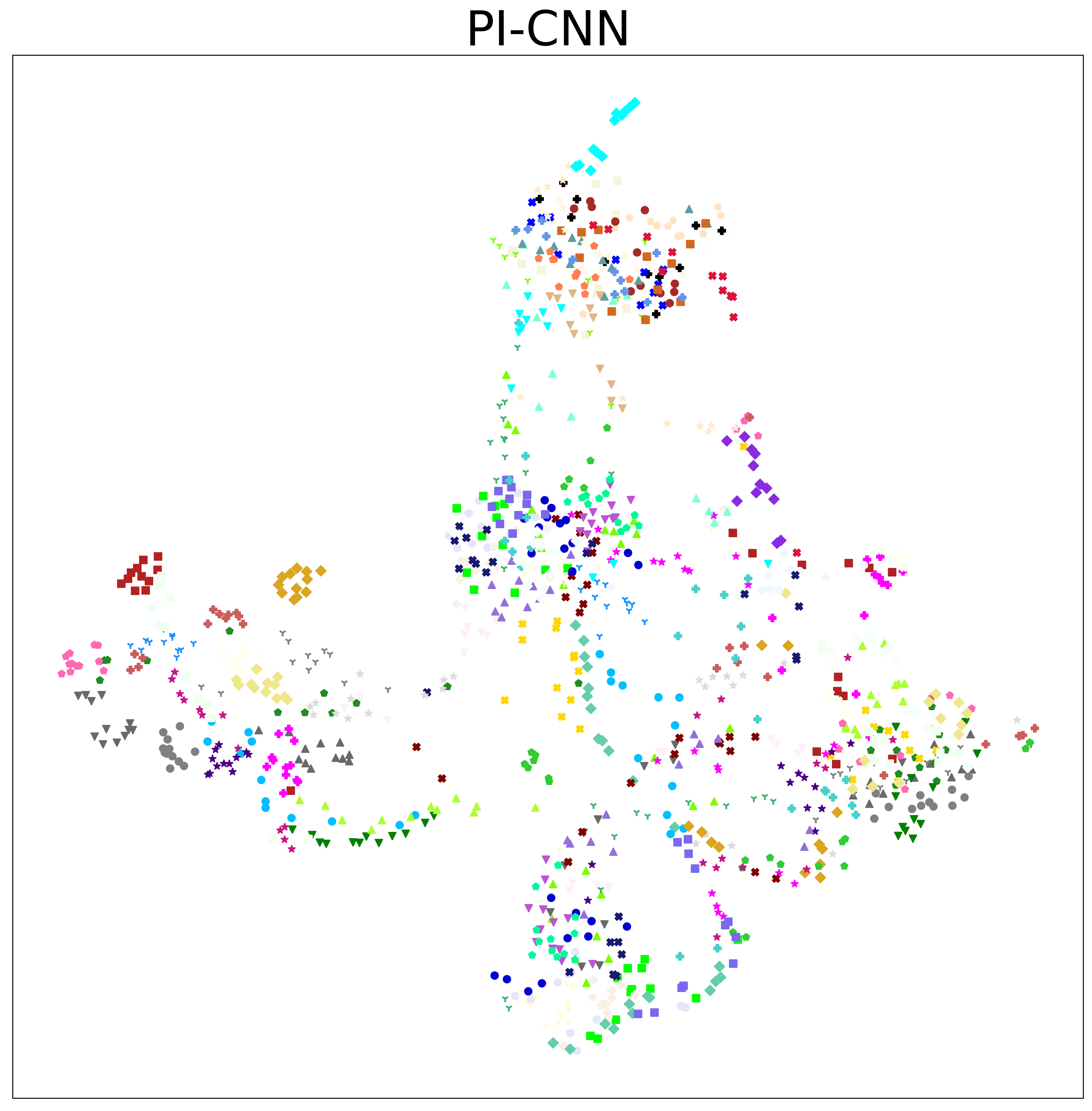}
  \hfill
		\includegraphics[height=0.2\textwidth, width=0.245\textwidth]{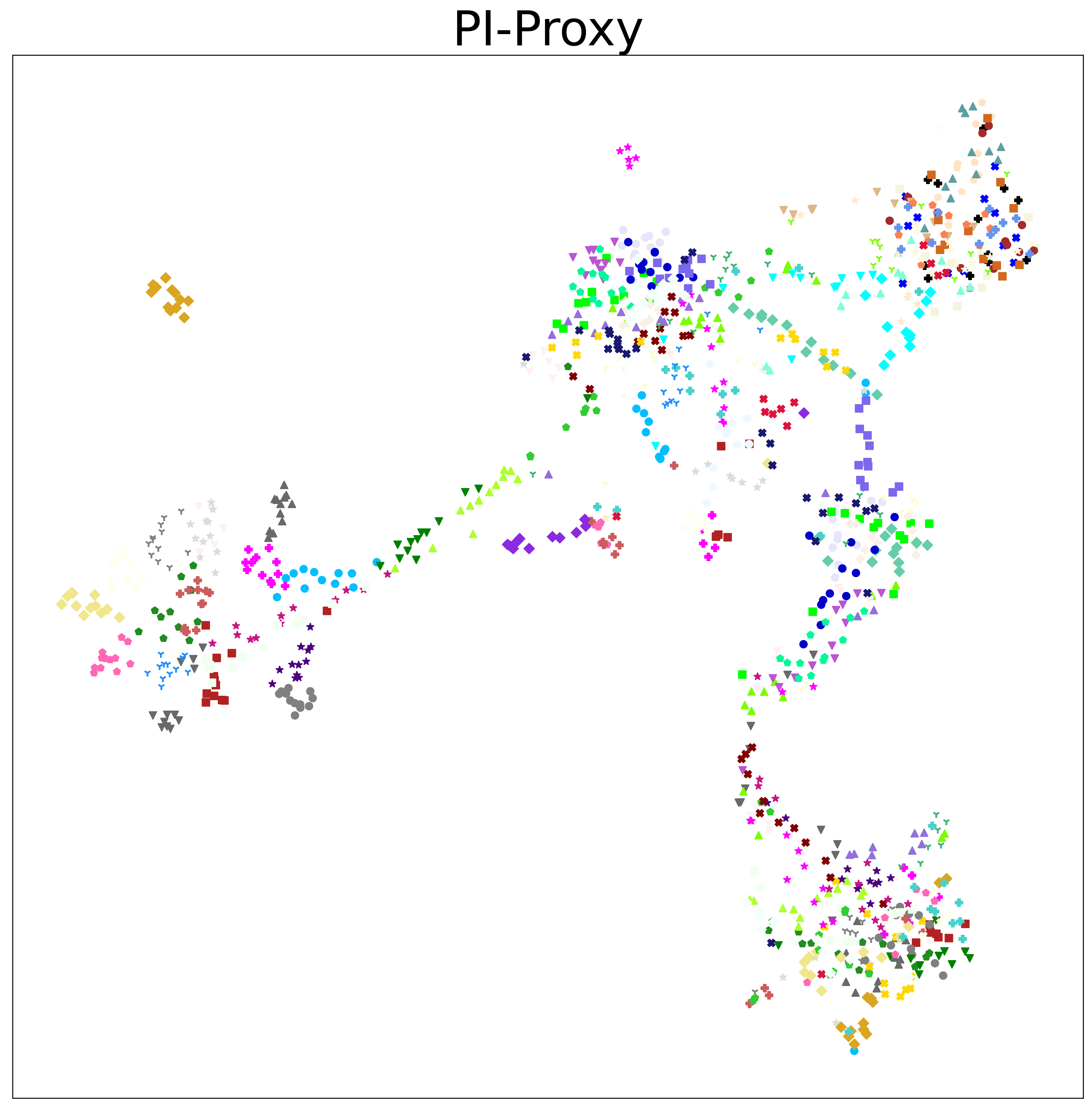}
  \hfill
  		\includegraphics[height=0.2\textwidth, width=0.245\textwidth]{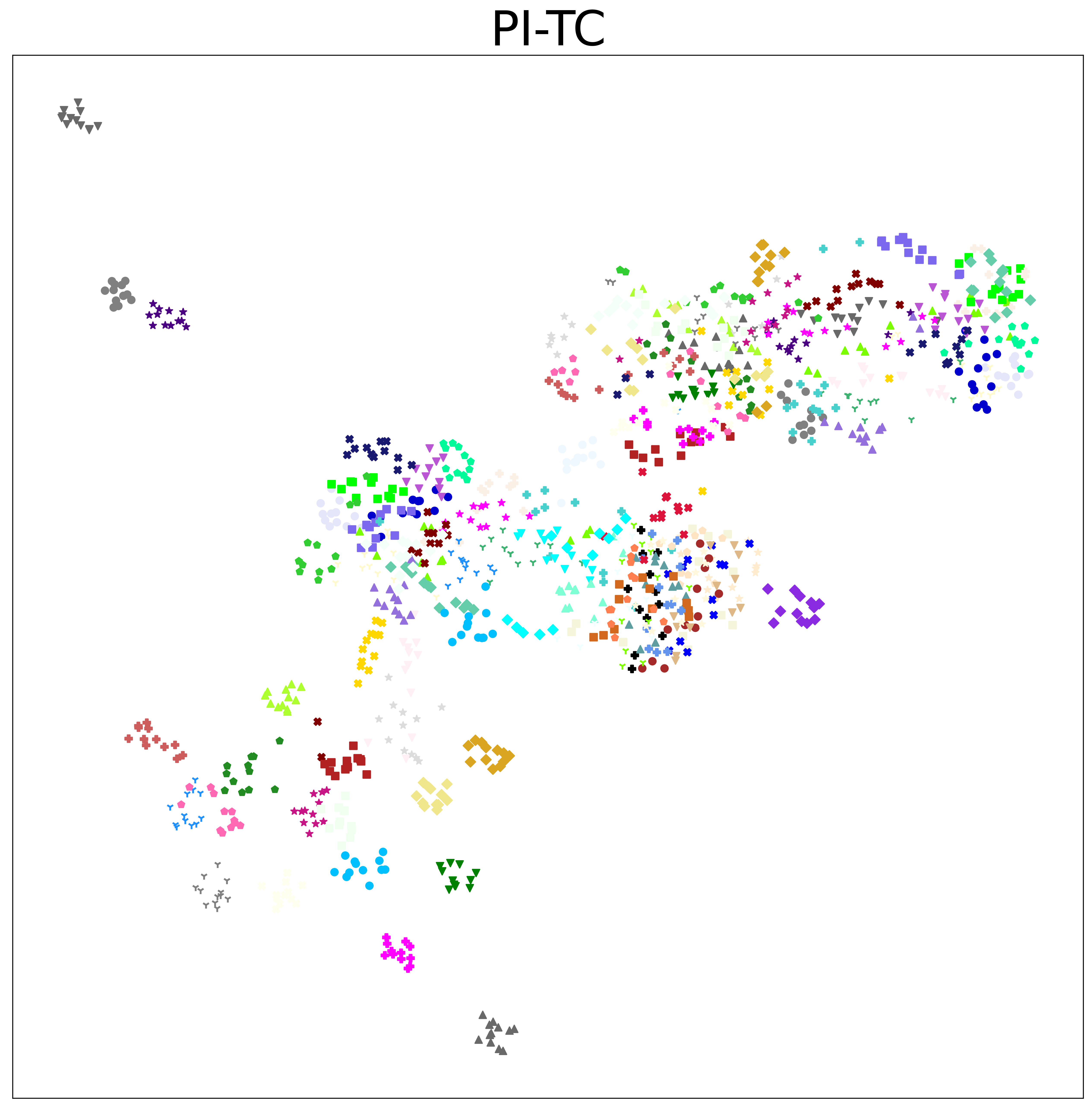}
    \hfill
		\includegraphics[height=0.2\textwidth, width=0.245\textwidth]{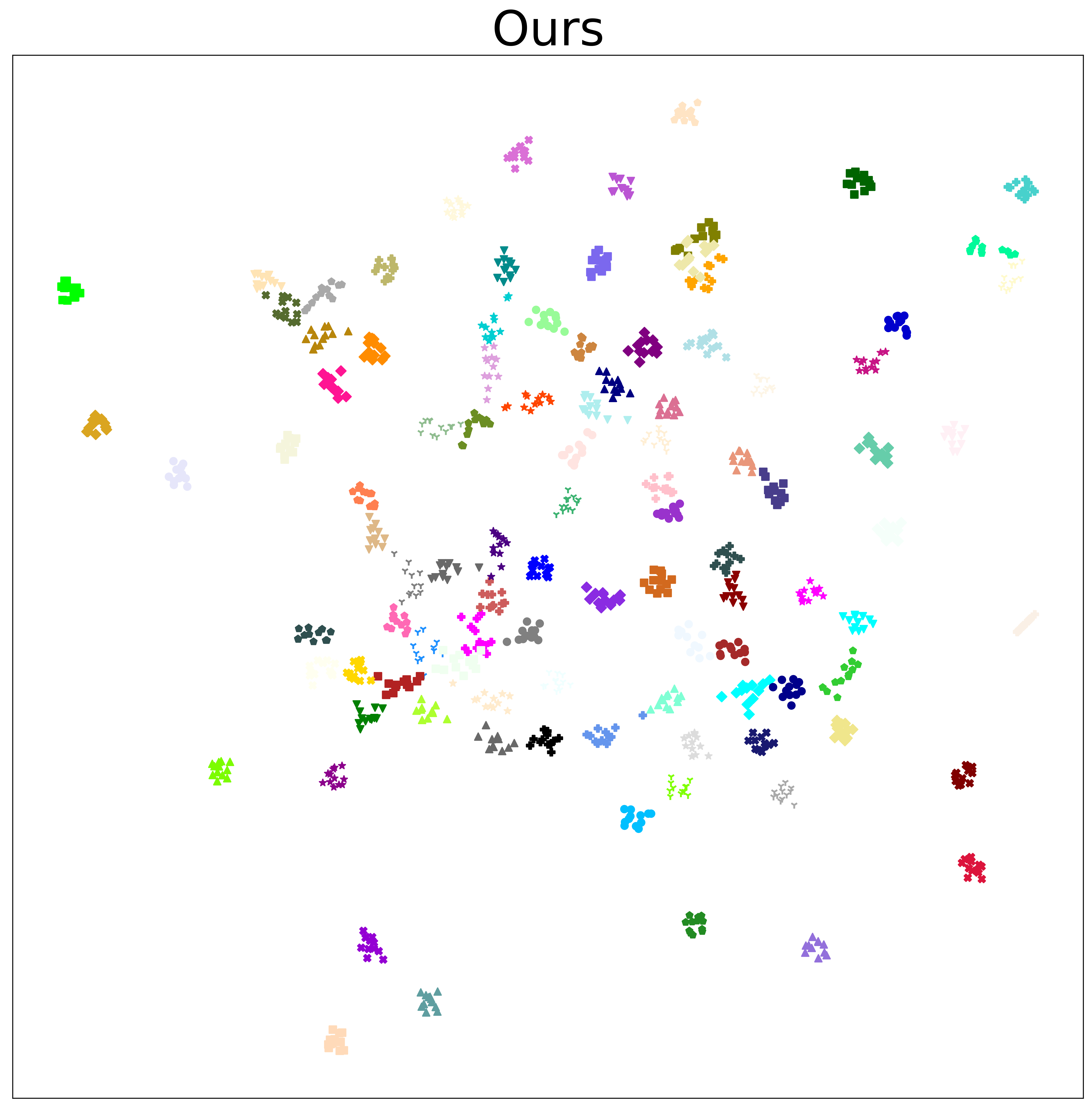}
	\end{minipage}
 \vspace{-0.05in}
	\caption{\em We show UMAP \cite{UMAP} visualizations for a qualitative comparison of the object embedding space learned for the ModelNet40 test dataset (from 5 categories such as table, desk, chair, stool, and sofa with 100 objects) by prior pose-invariant methods \cite{PIE2019} and our method. Each instance is an object view and a unique color and shape is used to denote each object-identity class in the visualizations. }
  \label{fig:viscomp}
  \vspace{-0.2in} 
\end{figure*}
\noindent 
In Table \ref{tbl:resultsPICR}, we compare performance of our method against several state-of-the-art multi-view and pose-invariant methods \cite{mvcnn15, PIE2019, TCL2018} reported by \cite{PIE2019} on the ModelNet-40 and ObjectPI datasets. For the single-view object recognition task, we report the results using the trained models provided by \cite{PIE2019}. 
As explained in Section \ref{sec:lit_review}(B), the multi-view methods are designed for category-based tasks when multiple images are available during inference. However, they perform poorly when only a single view is available and Pose-invariant (PI) methods outperform the multi-view (MV) methods on single-view tasks as they constrain the single-view embeddings to be clustered close to the multi-view embeddings. 
Although these pose-invariant methods encourage the clustering of different views of the same object, they don't effectively separate the confusing instances of neighboring objects from the same category in the embedding space. 
Hence, they don't capture discriminative features to distinguish between visually similar objects from the same category because of which they exhibit poor performance on the single-view object recognition and retrieval tasks. 

 We observe that our method \methodName-DE, outperforms the state-of-the-art methods on both the average classification (improvement of 7.7\% on ModelNet-40 and 2.6\% on ObjectPI) and retrieval tasks (improvement of 13.0\% on ModelNet-40 and 8.8\% on ObjectPI) when learning dual category and object embeddings. We notice a significant improvement in the single-view object recognition (accuracy improves by 20.0\% on ModelNet40 and 2.0\% on ObjectPI) and retrieval tasks (mAP improves by 33.7\% on ModelNet-40 and 18.8\% on ObjectPI). 
 Even in the single embedding space,  \methodName-SE shows improvements on object-based tasks compared to the state-of-the-art approaches. 
 \begin{table}
 \footnotesize
    \centering
 \begin{tabular}{@{}l|llllllll|@{}}
\toprule
\multirow{3}{*}{Method}  & \multicolumn{4}{l|}{Classification (Accuracy \%)} & \multicolumn{4}{c|}{Retrieval (mAP \%)} \\ \cmidrule(l){2-5}\cmidrule(l){6-9}
 & \multicolumn{1}{c}{\begin{tabular}[c]{@{}c@{}}SV\\ Cat\end{tabular}} & \multicolumn{1}{c}{\begin{tabular}[c]{@{}c@{}}MV\\ Cat\end{tabular}} & \multicolumn{1}{c}{\begin{tabular}[c]{@{}c@{}}SV\\ Obj\end{tabular}} & \multicolumn{1}{c|}{\bf Avg} & \multicolumn{1}{c}{\begin{tabular}[c]{@{}c@{}}SV\\ Cat\end{tabular}} & \multicolumn{1}{c}{\begin{tabular}[c]{@{}c@{}}MV\\ Cat\end{tabular}} & \multicolumn{1}{c}{\begin{tabular}[c]{@{}c@{}}SV\\ Obj\end{tabular}} & \multicolumn{1}{c|}{\bf Avg}\\ \midrule
PI-CNN & 79.7 & {\bf 83.3} & 23.6 & \multicolumn{1}{l|}{62.2} & 70.2 & 76.8 & 10.5 & 52.5 \\
PI-Proxy & {\bf 80.0} & 83.2 & 23.4 & \multicolumn{1}{l|}{62.2} & {\bf 70.6} & {\bf 77.0} & 10.7 & 52.8 \\
PI-TC & 76.1 & 82.5 & 36.5 & \multicolumn{1}{l|}{65.0} & 61.5 & 74.7 & 15.9 & 50.7 \\
\midrule
Ours & 78.9 & 81.8 & {\bf 83.0} & \multicolumn{1}{l|}{ \bf 81.2} & 68.0 & 74.2 & {\bf 72.8} & {\bf 71.7} \\
\bottomrule
\end{tabular}
\vspace{-0.05in} 
\caption{\em Comparison of performance on the FG3D dataset with state-of-the-art pose-invariant methods. }
\vspace{-0.25in}
\label{tbl:fg3d_comp}
\end{table}

 We train the state-of-the-art pose-invariant methods \cite{PIE2019} on the FG3D dataset and compare performance with our method in Table \ref{tbl:fg3d_comp}. FG3D is more challenging for object-level tasks as it comprises a large number of similar objects in each category with fine-grained differences. As mentioned earlier, prior methods mainly focus on
 learning category-specific embeddings and do not effectively separate the embeddings for objects within each category. In contrast, our proposed pose-invariant object loss separates confusing instances of objects from the same category which helps learn more discriminative fine-grained features to distinguish between visually similar objects resulting in significant improvement on single-view object recognition accuracy of 46.5\% and object retrieval mAP of 56.9\%. Overall, we outperform the pose-invariant methods on the classification tasks by 16.2\% and retrieval tasks by 18.9\%.

\noindent
{\bf (C) Ablation Studies: }\\
\label{sec:ablations}
\noindent
{\bf Visualization of pose-invariant embeddings: } From Fig. \ref{fig:viscomp}, we observe that for the pose-invariant methods (PI-CNN, PI-Proxy, and PI-TC), the embeddings for objects from the same category are not well-separated leading to poor performance on object-based tasks.
In contrast, the object embeddings generated using our method are much better separated as our pose-invariant object loss separates confusing instances of objects from the same category.  A more detailed comparison is shown in Supplemental Sec. \ref{sec:vis_pi}. \\
\begin{table*}[t]
\begin{minipage}{0.65\linewidth}
\setlength{\tabcolsep}{3.2pt}
\scriptsize
\begin{tabular}{@{}lclllllllllll@{}}
\toprule
\multirow{3}{*}{Dataset} & \multirow{4}{*}{\begin{tabular}[c]{@{}c@{}}Embed.\\ Space\end{tabular}} & \multicolumn{1}{c}{\multirow{3}{*}{Losses}} & \multicolumn{5}{c}{Classification (Accuracy \%)} & \multicolumn{5}{c}{Retrieval (mAP \%)} \\ \cmidrule(l){4-8} \cmidrule(lr){9-13} 
 &  & \multicolumn{1}{c}{} & \multicolumn{2}{c}{Category} & \multicolumn{2}{c}{Object} & \multicolumn{1}{c}{\multirow{2}{*}{Avg.}} & \multicolumn{2}{c}{Category} & \multicolumn{2}{c}{Object} & \multicolumn{1}{c}{\multirow{2}{*}{Avg.}} \\ \cmidrule(lr){4-5} \cmidrule(lr){6-7} \cmidrule(lr){9-10} \cmidrule(lr){11-12}
 &  & \multicolumn{1}{c}{} & \multicolumn{1}{c}{SV} & \multicolumn{1}{c}{MV} & \multicolumn{1}{c}{SV} & \multicolumn{1}{c}{MV} & \multicolumn{1}{c}{} & \multicolumn{1}{c}{SV} & \multicolumn{1}{c}{MV} & \multicolumn{1}{c}{SV} & \multicolumn{1}{c}{MV} & \multicolumn{1}{c}{} \\ \midrule 
\multirow{6}{*}{Object} & \multirow{3}{*}{Single} & {\tiny $\mathcal{L}_{cat}$} & 70.7 & 81.6 & 78.7 & 87.8 & 79.7 & 65.3 & 82.9 & 54.8 & 92.9 & 73.9 \\
\multirow{6}{*}{$\quad$ PI}  &  & {\tiny $\mathcal{L}_{cat}+\mathcal{L}_{piobj}$} & 69.4 & 81.6 & 88.5 & 98.0 & 84.4 & {\bf 66.0} & 75.6 & 68.5 & 98.9 & 77.2 \\ 
 &  & {\tiny $\mathcal{L}_{cat} + \mathcal{L}_{piobj} + \mathcal{L}_{picat}$} & 71.2 & 82.7 & 83.3 & 95.9 & 83.3 & 65.6 & 82.8 & 62.3 & 98.0 & 77.2 \\ \cmidrule(l){2-13} 
 & \multirow{2}{*}{Dual} & {\tiny $\mathcal{L}_{cat} + \mathcal{L}_{piobj}$} & 71.2 & 82.7 & \textbf{94.5} & \textbf{99.0} & \textbf{86.8} & 65.7 & 82.9 & 80.5 & \textbf{99.5} & 82.2 \\
 &  & {\tiny $\mathcal{L}_{cat} + \mathcal{L}_{piobj} + \mathcal{L}_{picat}$} & \textbf{71.3} & \textbf{83.7} & 92.7 & 98.0 & 86.4 & 65.7 & \textbf{83.4} & \textbf{81.0} & 99.0 & \textbf{82.3} \\ \midrule
\multirow{6}{*}{Model} & \multirow{3}{*}{Single} & {\tiny $\mathcal{L}_{cat}$} & 84.7 & 88.4 & 71.3 & 75.9 & 80.1 & 79.0 & 84.8 & 45.3 & 82.0 & 72.8 \\
\multirow{6}{*}{Net40} &  & {\tiny $\mathcal{L}_{cat} + \mathcal{L}_{piobj}$} & \textbf{85.4} & 88.8 & 81.2 & 85.6 & 85.2 & 79.1 & 83.1 & 59.2 & 90.4 & 78.0 \\
 &  & {\tiny $\mathcal{L}_{cat} + \mathcal{L}_{piobj} + \mathcal{L}_{picat}$} & 84.7 & 88.4 & 71.8 & 79.3 & 81.0 & 78.7 & 84.9 & 49.1 & 85.2 & 74.5 \\ \cmidrule(l){2-13} 
 & \multirow{2}{*}{Dual} & {\tiny $\mathcal{L}_{cat} + \mathcal{L}_{piobj}$} & 84.5 & 88.6 & \textbf{94.6} & 96.6 & 91.1 & 78.9 & 85.0 & \textbf{85.2} & 98.1 & 86.8 \\
 &  & {\tiny $\mathcal{L}_{cat} + \mathcal{L}_{piobj} + \mathcal{L}_{picat}$} & 85.2 & \textbf{88.9} & 93.7 & \textbf{96.9} & \textbf{91.2} & \textbf{79.7} & \textbf{86.1} & 84.0 & \textbf{98.2} & \textbf{87.0} \\ 
 \midrule
\multirow{6}{*}{FG3D} & \multirow{3}{*}{Single} & {\tiny $\mathcal{L}_{cat}$} & {\bf 79.3} & 81.8 & 18.2 & 19.0 & 49.5 & 66.6 & 73.1 & 9.7 & 28.4 & 44.5 \\
\multirow{6}{*}{} &  & {\tiny $\mathcal{L}_{cat} + \mathcal{L}_{piobj}$} & 78.3 & 80.2 & 26.2 & 31.0 & 53.9 & 64.9 & 69.0 & 15.7 & 42.9 & 48.1 \\
 &  & {\tiny $\mathcal{L}_{cat} + \mathcal{L}_{piobj} + \mathcal{L}_{picat}$} & 78.4 & 81.1 & 29.3 & 41.8 & 57.6 & 65.1 & 70.8 & 17.9 & 55.0 & 52.2 \\ \cmidrule(l){2-13} 
 & \multirow{2}{*}{Dual} & {\tiny $\mathcal{L}_{cat} + \mathcal{L}_{piobj}$} & 78.7 & {\bf 82.2} & {\bf 83.2} & 91.4 & {\bf 83.9} & 67.6 & 73.1 & 72.8 & 95.3 & 77.2\\
 &  & {\tiny $\mathcal{L}_{cat} + \mathcal{L}_{piobj} + \mathcal{L}_{picat}$} & 79.0 & 81.9 & 83.1 & \textbf{91.6} & \textbf{83.9} & \textbf{68.1} & \textbf{74.4} & {\bf 73.0} & \textbf{95.5} & \textbf{77.8} \\
 \bottomrule
\end{tabular}
\vspace{-0.05in} 
\caption{\em Ablations of the proposed losses in the single and dual embedding spaces. }
\label{tbl:abl_dual_piloss}
\vspace{-0.1in}
\hfill
\vline 
\hfill 
\end{minipage}
\begin{minipage}{0.34\linewidth}
    \includegraphics[height=0.96\textwidth, width=\textwidth]{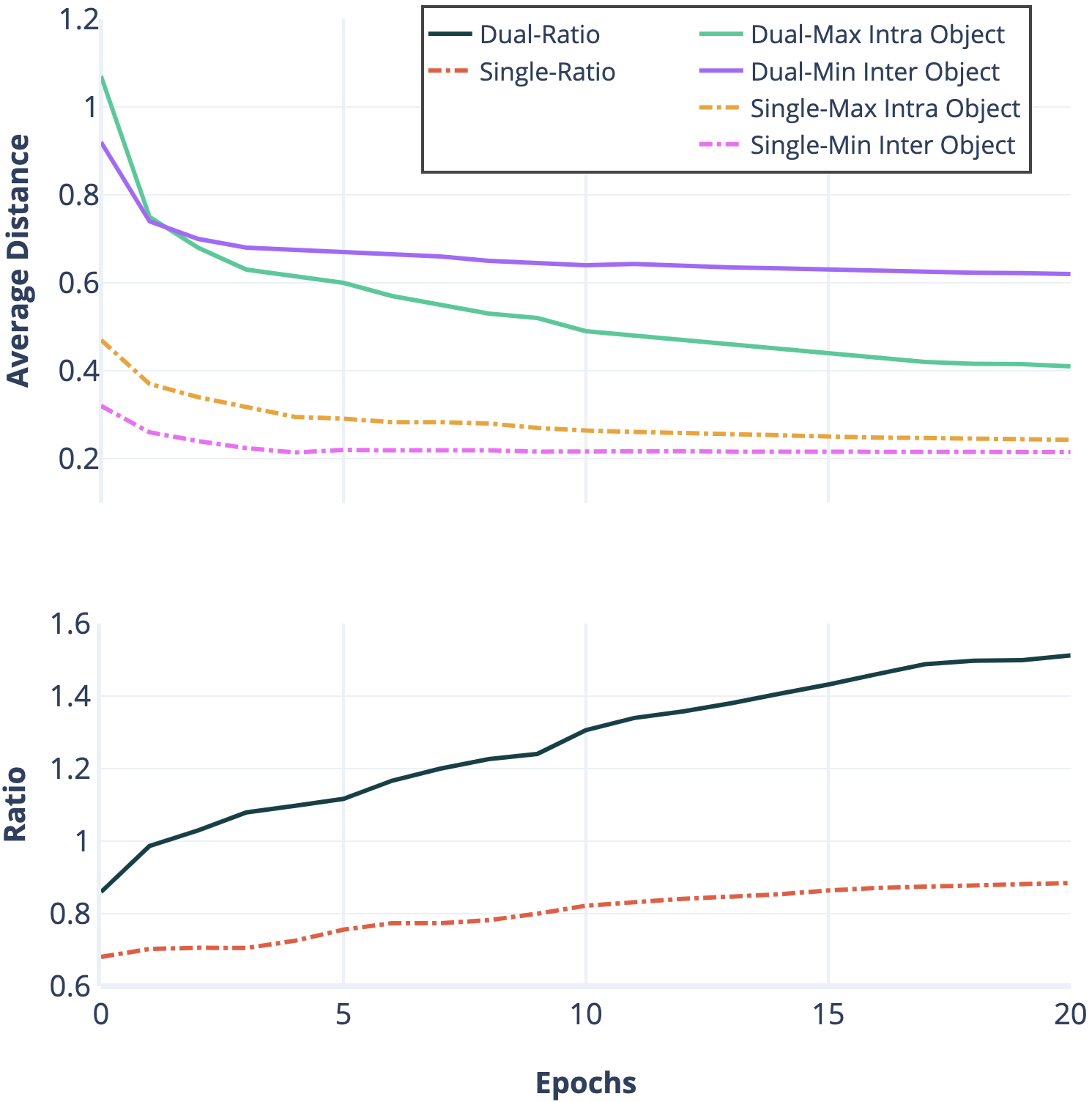}
    \vspace{-0.28in}
\captionof{figure}{\em Optimization of the inter-class and intra-class distances for object-identity classes during training while learning single and dual embedding spaces for the ModelNet40 dataset. }
\label{fig:dists}

\end{minipage}
\vspace{-0.2in}
\end{table*}
\noindent
{\bf Single and dual embedding spaces: }
From Tables \ref{tbl:single_dual} and \ref{tbl:abl_dual_piloss}, we observe that learning dual embeddings leads to better overall performance, especially for object-based tasks. This is because, for category-based tasks, we aim to embed objects from the same category close to each other while for object-based tasks, we aim to separate objects apart from each other to be able to discriminate between them. This leads to contradicting goals for object and category-based tasks in the single embedding space. 
Learning dual embeddings more effectively captures category and object-specific attributes in separate representation spaces leading to overall performance improvements. \\
\noindent
{\bf Pose-invariant losses: }
We employ three losses in \methodName{}: $\mathcal{L}_{cat}$ to distinguish between different categories, $\mathcal{L}_{picat}$ for clustering objects from the same category, and $\mathcal{L}_{piobj}$ for clustering different views of the same object and separating confusing instances from different objects of the same category for object-based tasks. Table \ref{tbl:abl_dual_piloss} shows that in the single embedding space, $\mathcal{L}_{cat}$ is effective for category-based tasks, but not for object-based tasks. Adding $\mathcal{L}_{piobj}$ improves performance in object-based tasks, but at the cost of category-based tasks (especially MV category retrieval). This can be mitigated by adding $\mathcal{L}_{picat}$ that enhances performance on category-based tasks. However, $\mathcal{L}_{picat}$ and $\mathcal{L}_{piobj}$ have conflicting objectives in the same space and only marginally improve overall performance over $\mathcal{L}_{cat}$ in the single embedding space.
In the dual embedding space, these losses are optimized in separate embedding spaces. In the dual space, we observe that $\mathcal{L}_{cat}+\mathcal{L}_{piobj}$ improves overall performance, particularly for object-based tasks, and adding $\mathcal{L}_{picat}$ boosts performance on category-based tasks and yields the best overall performance for all the datasets.
$\mathcal{L}_{piobj}$ enhances the separability of object-identity classes facilitating learning more discriminative object embeddings that
significantly improves performance on object-based tasks (see detailed ablation study of $\mathcal{L}_{piobj}$ in Sup. Sec \ref{sec:abl_piobj}).
\\
\begin{figure}[b]
	\centering
       \begin{minipage}{0.49\linewidth}
   \includegraphics[width=\textwidth]{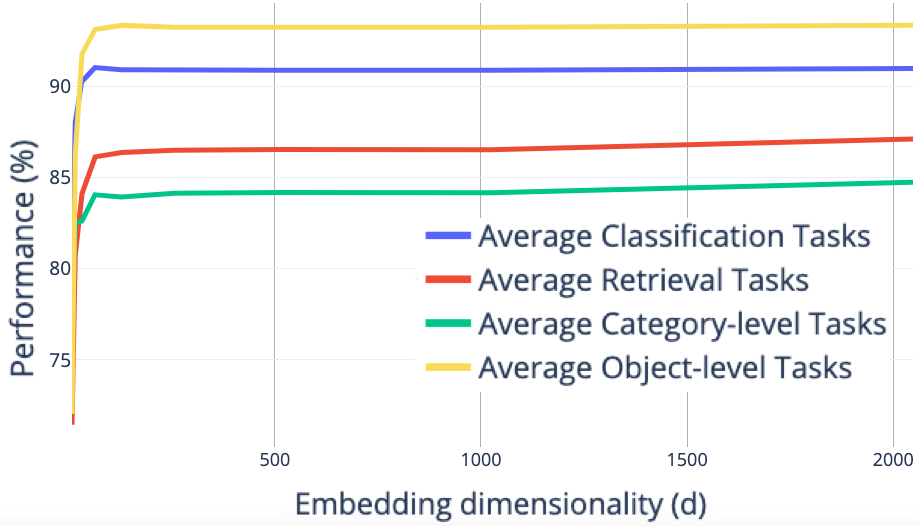}
   \vspace{-0.25in}
       \caption*{\em (a) ModelNet-40 dataset}
       \end{minipage}
		\begin{minipage}{0.49\linewidth}
		\includegraphics[width=\textwidth]{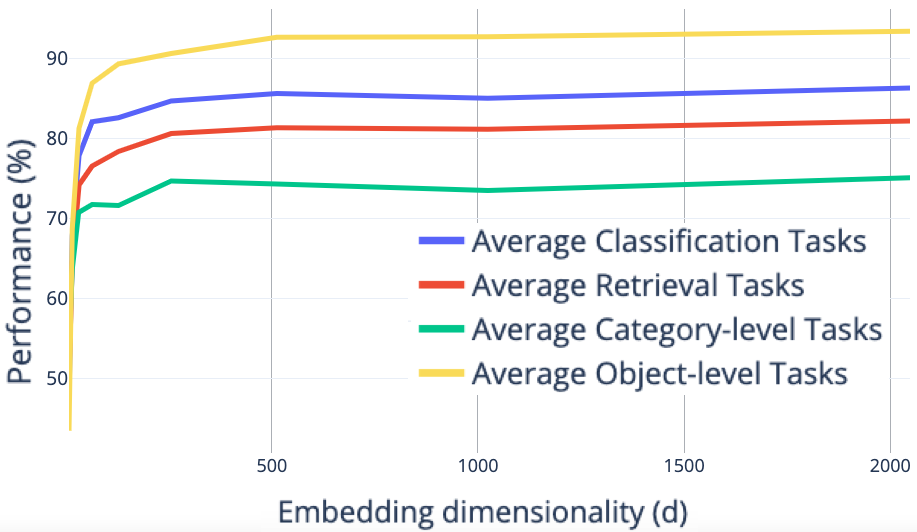}
       \vspace{-0.25in}
       \caption*{\em (b) ObjectPI dataset}
		\end{minipage}
   \vspace{-0.1in}
	\caption{\em Effect of embedding dimensionality on performance. 
\vspace{-0.1in}
}
\label{fig:vis_embed_dim}
\end{figure}
\noindent 
{\bf Optimizing intra-class and inter-class distances: } In the top of Fig. \ref{fig:dists}, we show the maximum intra-class distance ($d^{max}_{intra}$) and minimum inter-class distance ($d^{min}_{inter}$) between object-identity classes from the same category during training on the ModelNet40 dataset. 
These distances are computed using the object-identity embeddings and averaged over all objects. 
We also plot the ratio $\rho=\frac{d^{min}_{inter}}{d^{max}_{intra}}$ in the bottom of Fig. \ref{fig:dists}. A higher $\rho$ value indicates embeddings of the same object-identity class are well clustered and separated from embeddings of other object-identity classes from the same category. Comparing the plots for the single and dual embedding spaces, we observe that $\rho$ and $d^{min}_{inter}$ are much higher for the dual space indicating better separability of object-identity classes when learning a dual space. We observe the same effect for all datasets %\ref{sec:opt_dist}).} 
(see Sup. Sec \ref{sec:abl_opt_dist}).
\\
\noindent 
{\bf Embedding dimensionality: } In Fig. \ref{fig:vis_embed_dim},
 we observe that for ModelNet-40, a dimension of 64 for category and 128 for object-level tasks is sufficient for good performance. For ObjectPI, higher dimensions of 256 and 512 are required for category and object-level tasks respectively to capture color and texture information in addition to shape, unlike ModelNet-40. A higher embedding dimensionality is required for object-level tasks compared to category-level tasks possibly because object embeddings need to capture finer details to effectively distinguish between objects.
We provide more details in the Supplemental Sec. \ref{sec:emb_dim}.

\noindent
{\bf Qualitative results: } 
In the Supplemental, we illustrate how self-attention captures correlations between different views of an object using multi-view attention maps in Sec. \ref{sec:mv_maps},
 and present qualitative object retrieval results in Sec. \ref{sec:retr_vis}.
\section{Conclusion}
\label{sec:conclusion}
\noindent 
We propose a multi-view dual-encoder architecture and pose-invariant ranking losses that facilitate learning discriminative pose-invariant representations for joint category and object recognition and retrieval. Our method outperforms state-of-the-art methods on several pose-invariant classification and retrieval tasks on three publicly available multi-view object datasets. We further provide ablation studies to demonstrate the effectiveness of our approach.

%\clearpage
{
    \small
    \bibliographystyle{ieeenat_fullname.bst}
    \bibliography{main}
}
\clearpage
\maketitlesupplementary

\begin{figure*}[b]
	\centering
   
		\centering
		\fbox{\includegraphics[height=0.24\textwidth]{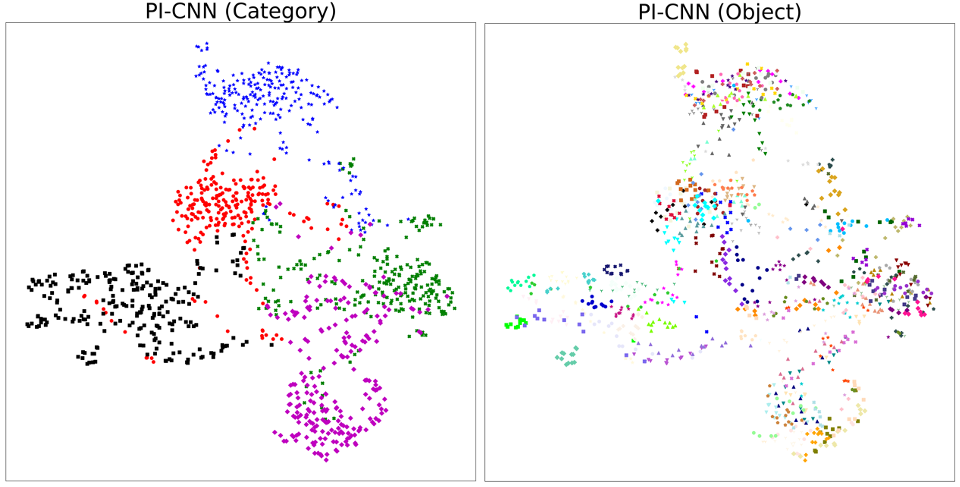}}
		\fbox{\includegraphics[height=0.24\textwidth]{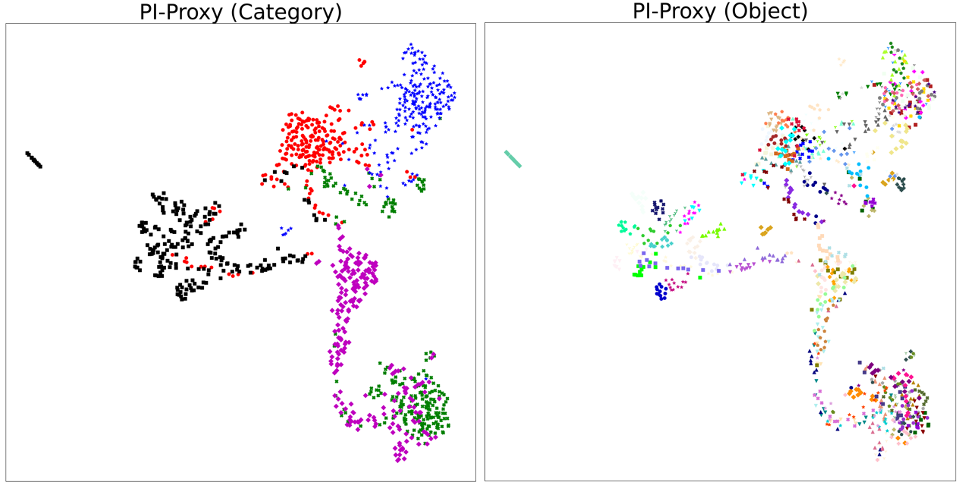}}
		\fbox{\includegraphics[height=0.24\textwidth]{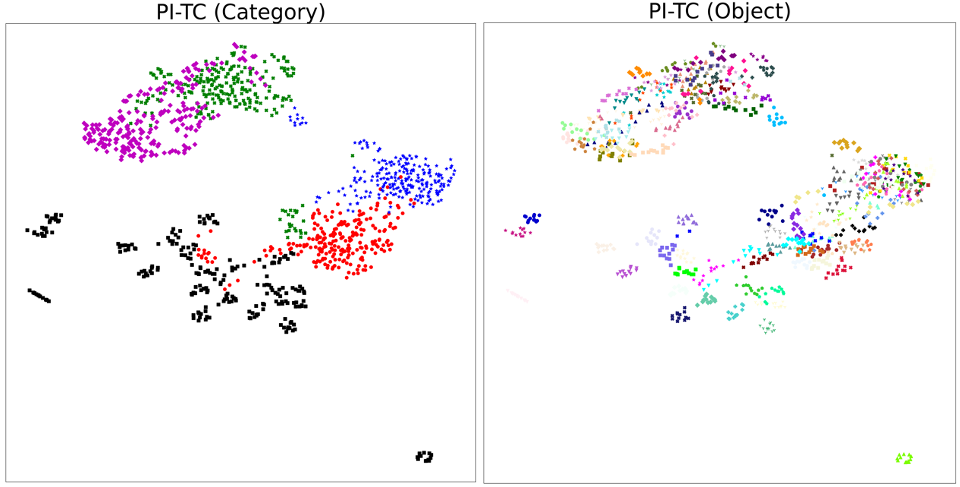}}
		\fbox{\includegraphics[height=0.24\textwidth]{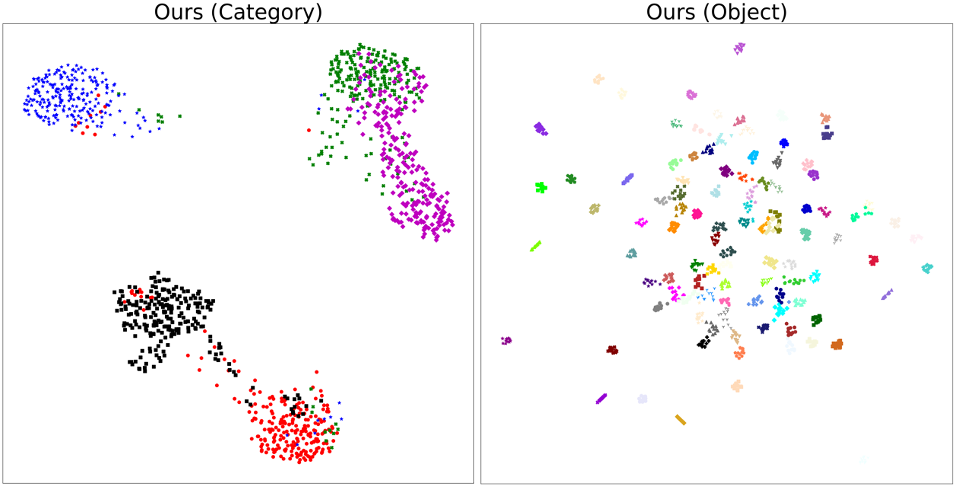}}
%    \caption*{(b) UMAP visualization}
	\caption{\em Qualitative comparison of the embedding space learned for a subset of the ModelNet40 test dataset (from 5 categories such as table, desk, chair, stool, sofa with 100 objects) by prior pose-invariant methods \cite{PIE2019} and our method (bottom-right). In the category plots (to the left of each subfigure), we use five distinct colors to indicate instances from each of the categories. In the object plots (to the right of each subfigure), instances of the same object-identity class are indicated by a unique color and shape. 
 }
\label{fig:umap_mnet40}
\end{figure*}

\begin{figure*}[h]
	\centering
		\includegraphics[height=0.25\textwidth]{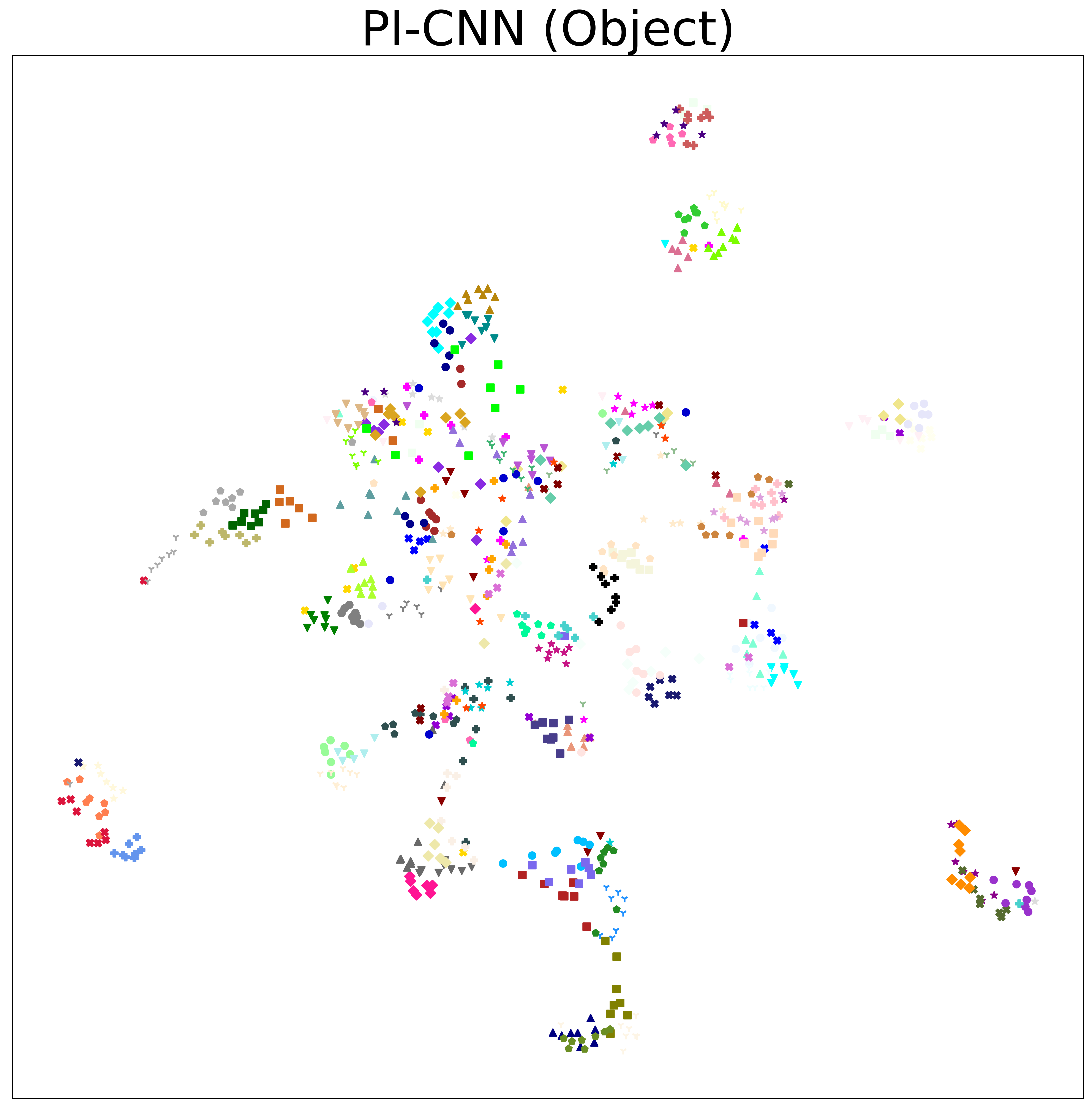}
		\includegraphics[height=0.25\textwidth]{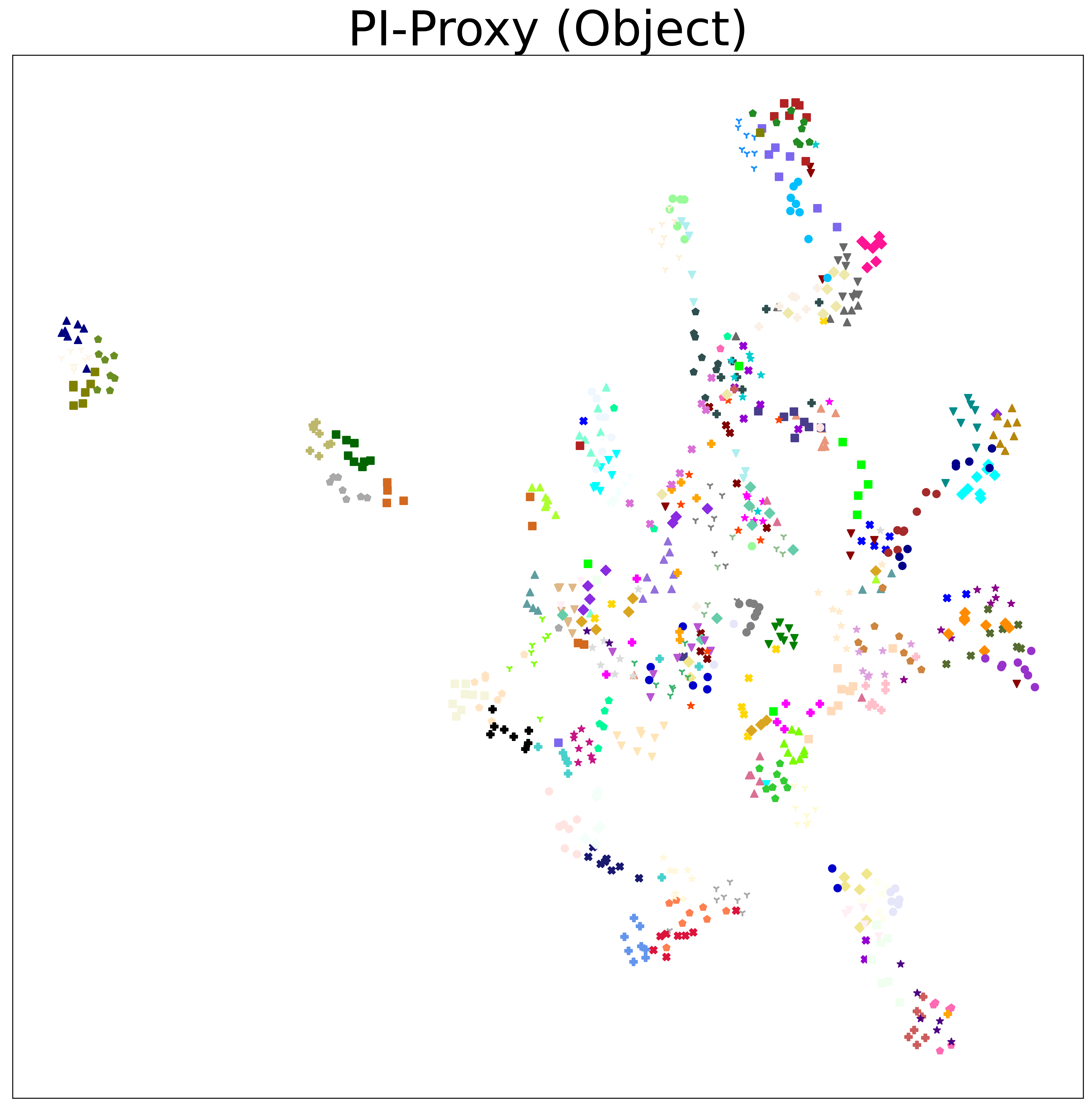}
		\includegraphics[height=0.25\textwidth]{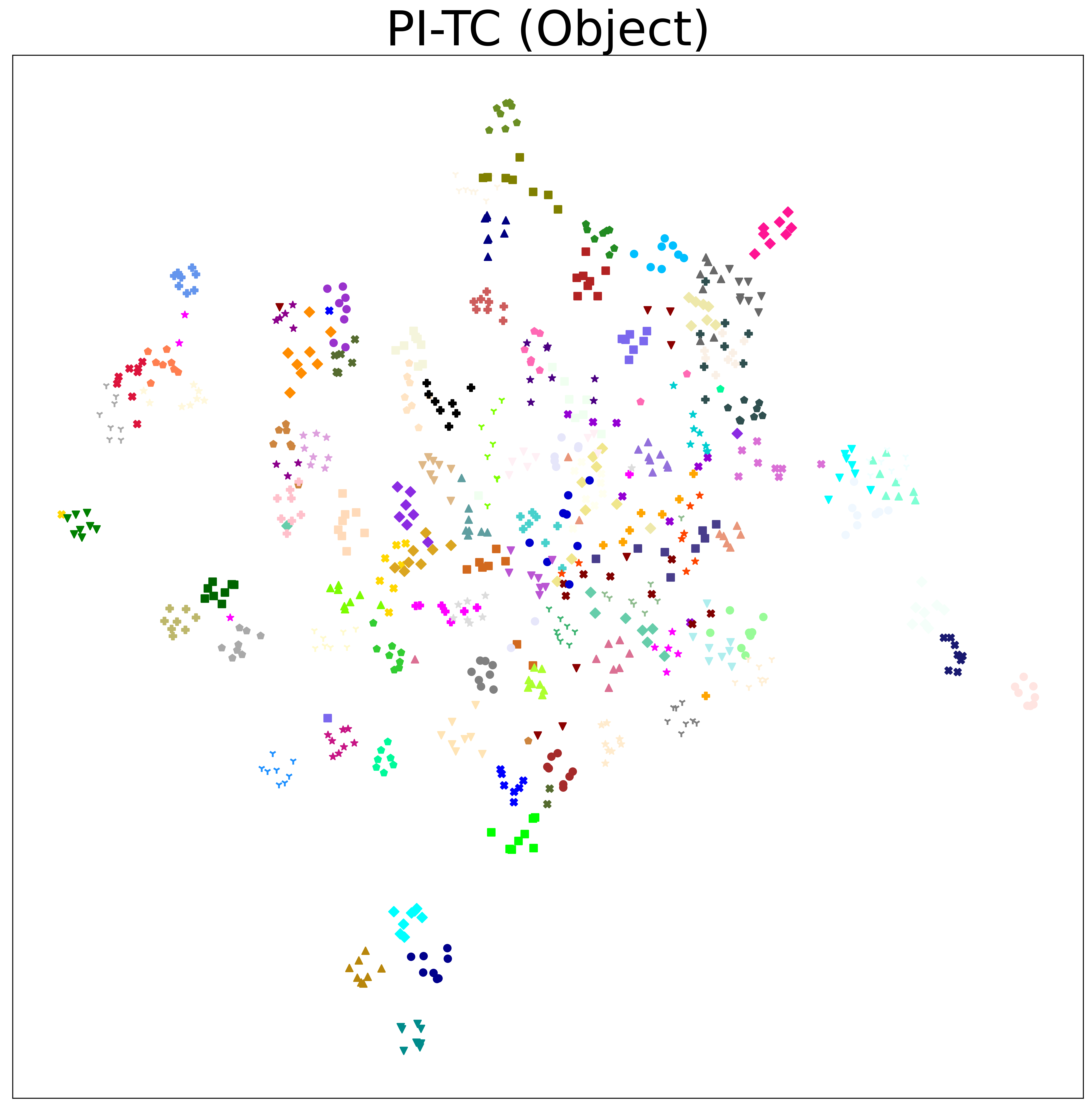}
		\includegraphics[height=0.25\textwidth]{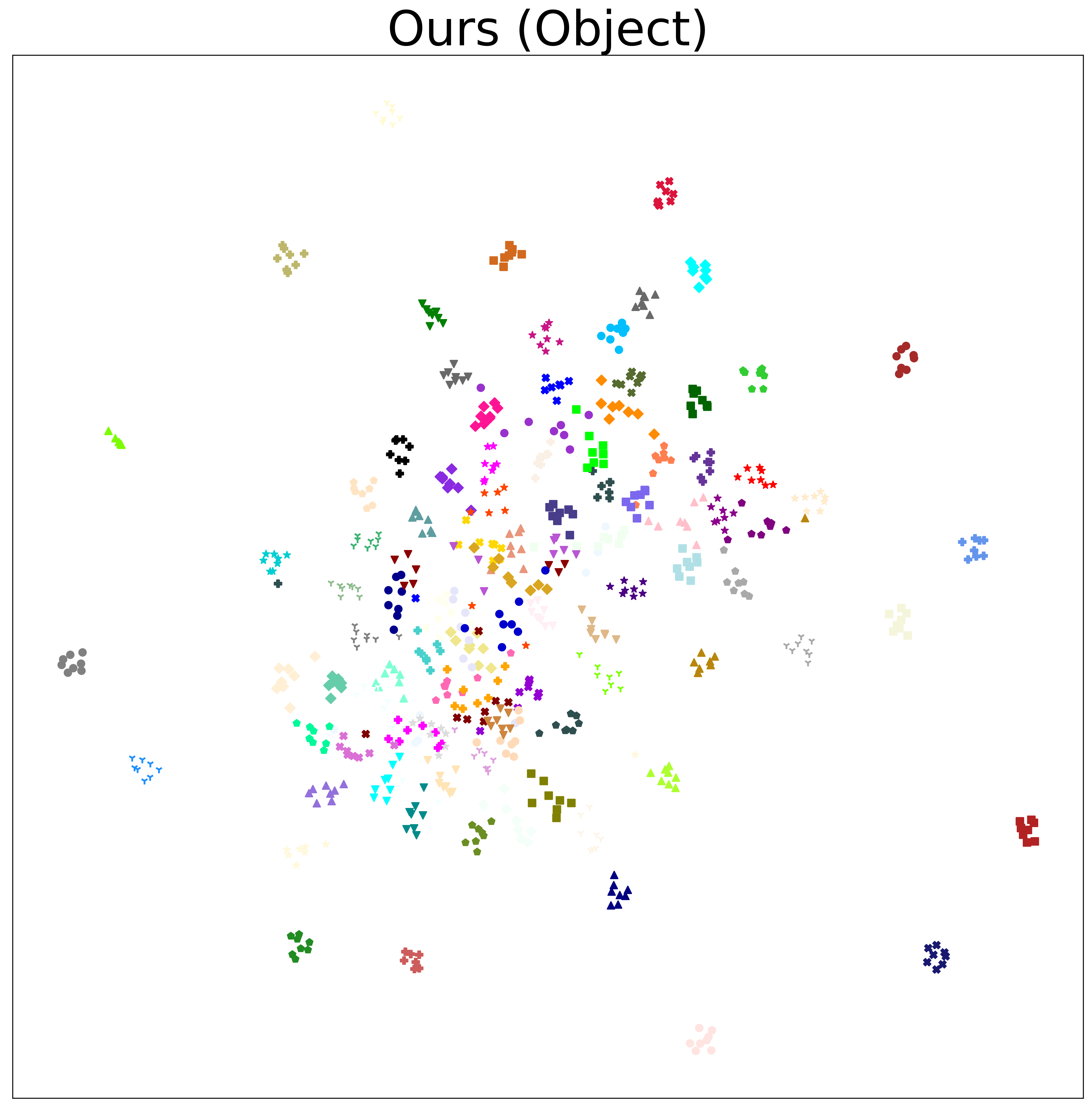}
	%\caption*{\em (b) UMAP visualization}
 \vspace{-0.25in}
 \caption{\em Comparison of the object embedding space learned for the ObjectPI test dataset (with 98 objects) by prior pose-invariant methods \cite{PIE2019} and our method (right). Each instance is an object view and each object-identity class is denoted by a unique color and shape. 
 }
\label{fig:umap_objectpi}
\vspace{0.15in}
		\includegraphics[height=0.25\textwidth]{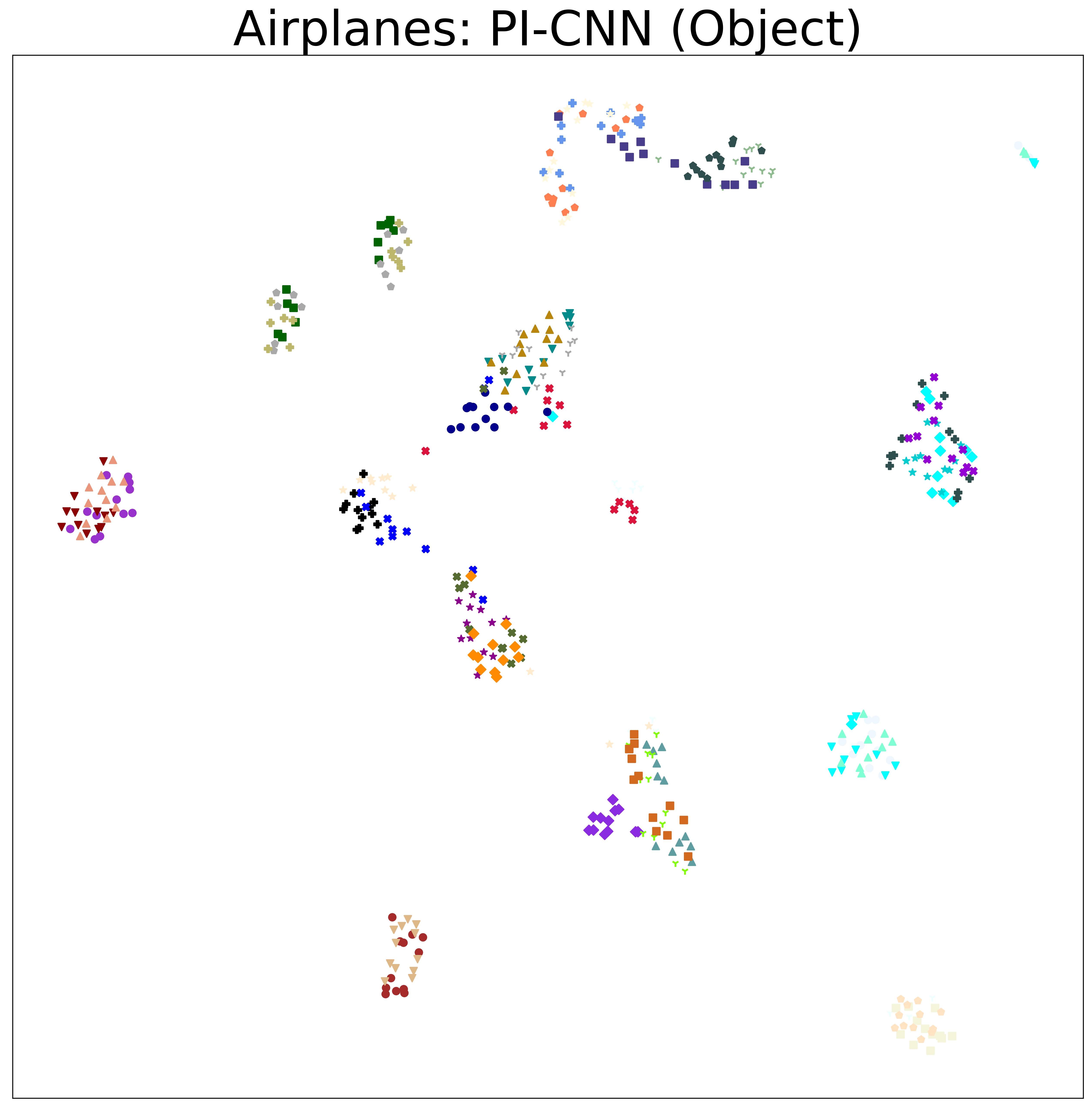}
		\includegraphics[height=0.25\textwidth]{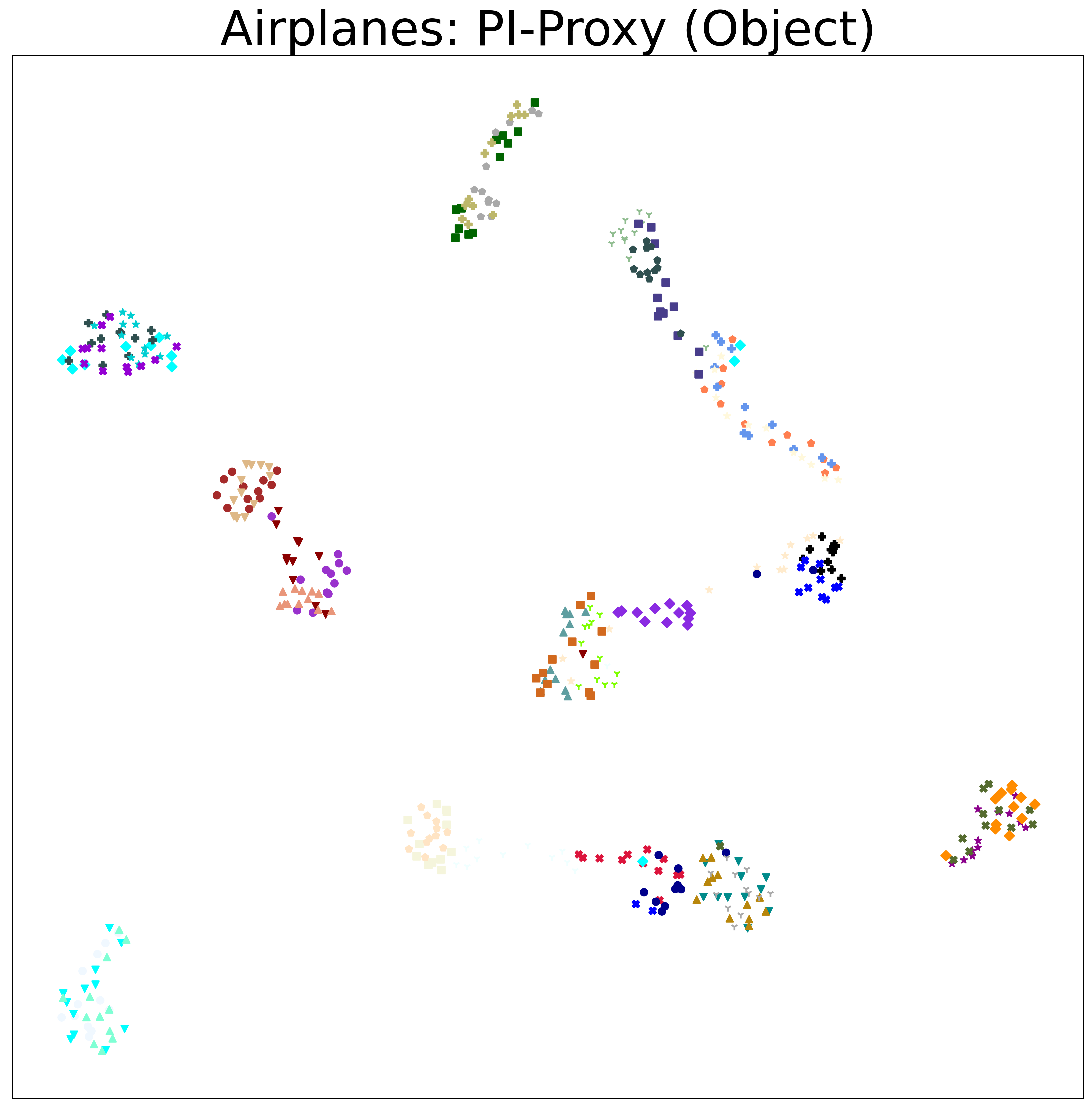}
		\includegraphics[height=0.25\textwidth]{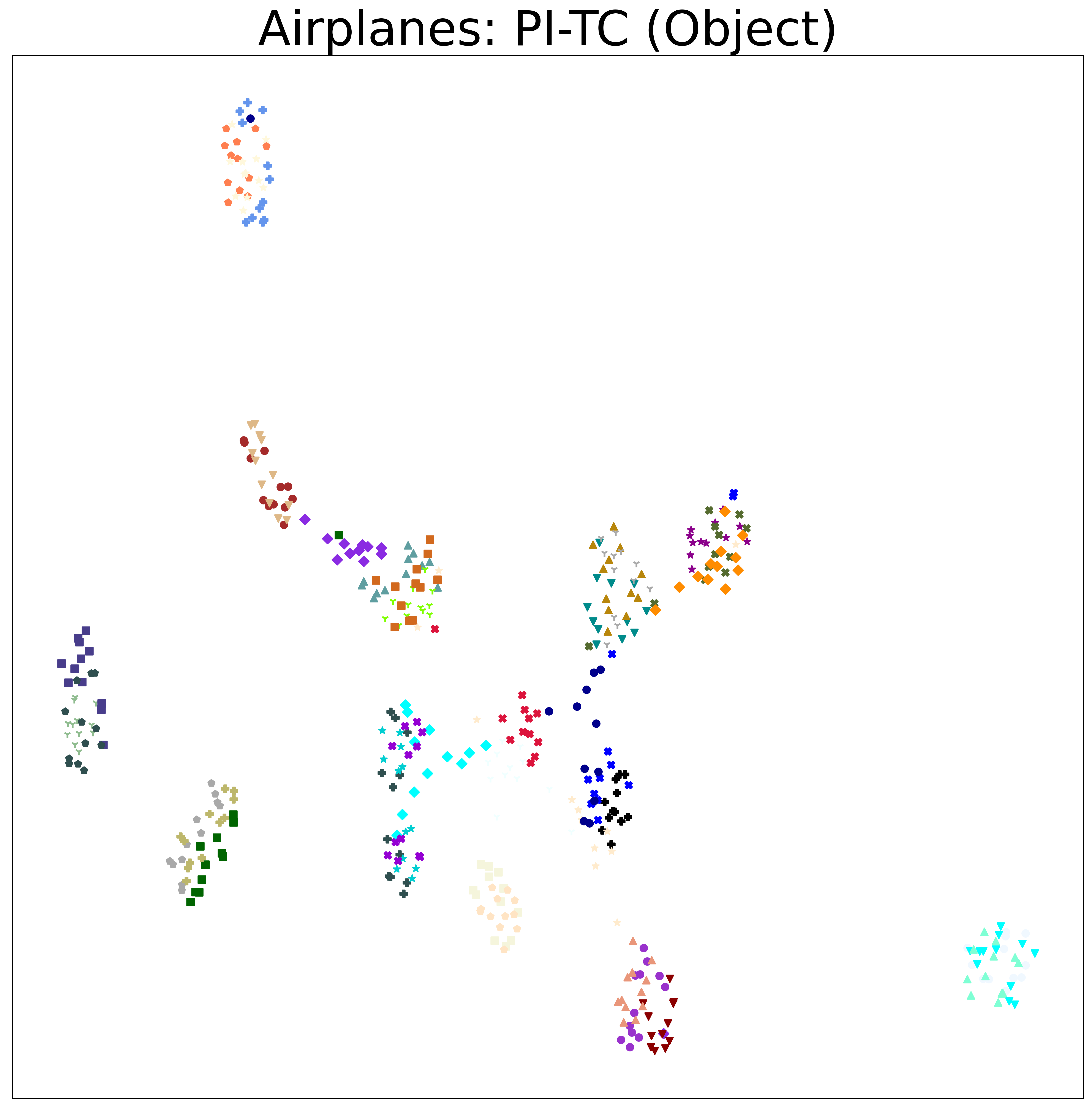}
		\includegraphics[height=0.25\textwidth]{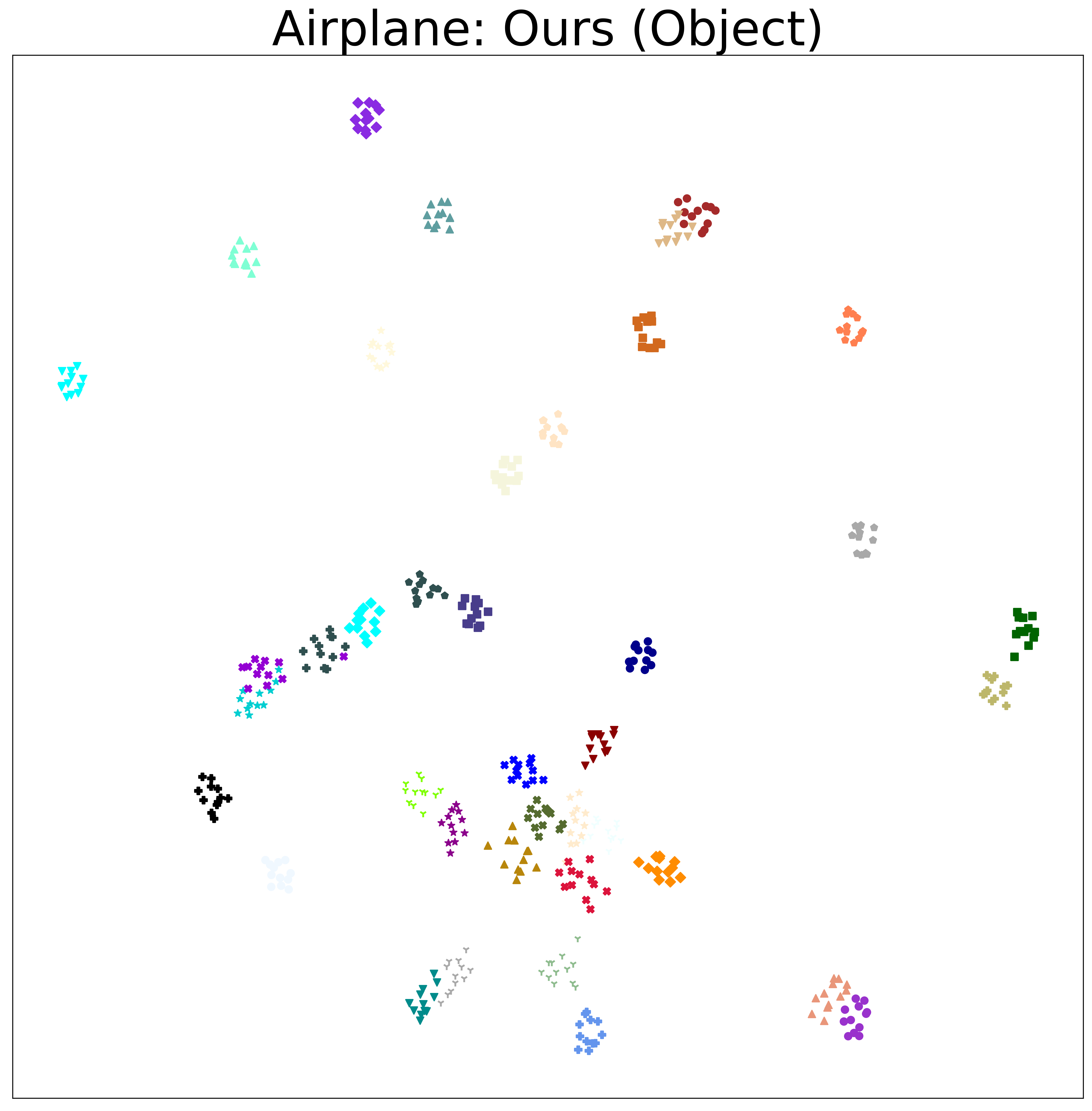}
	%\caption*{\em (b) UMAP visualization}
 \vspace{-0.25in}
 \caption*{\em (a) Object-identity embeddings for 39 airplane objects (3 objects each from 13 airplane categories)}
		\includegraphics[height=0.25\textwidth]{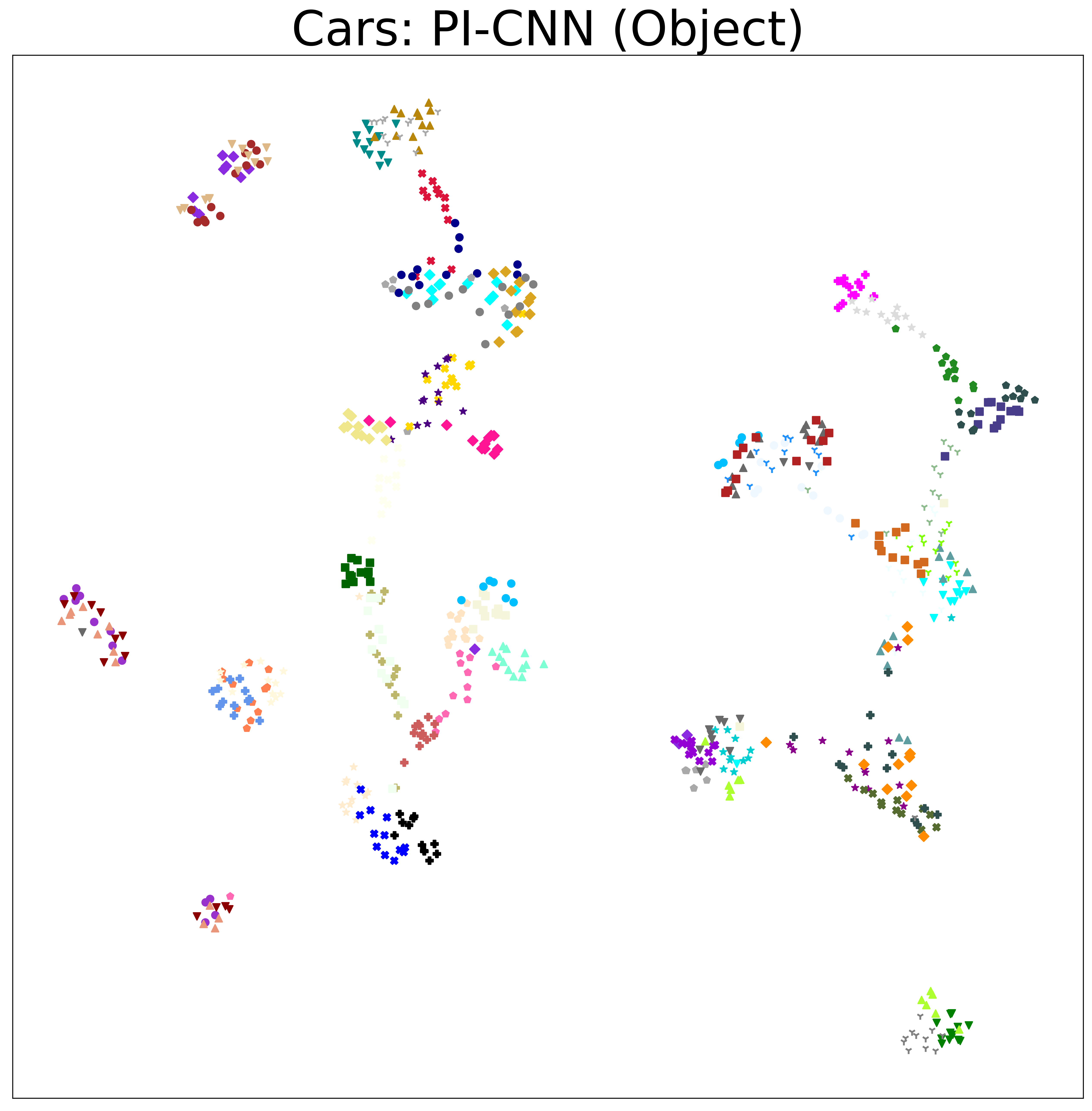}
		\includegraphics[height=0.25\textwidth]{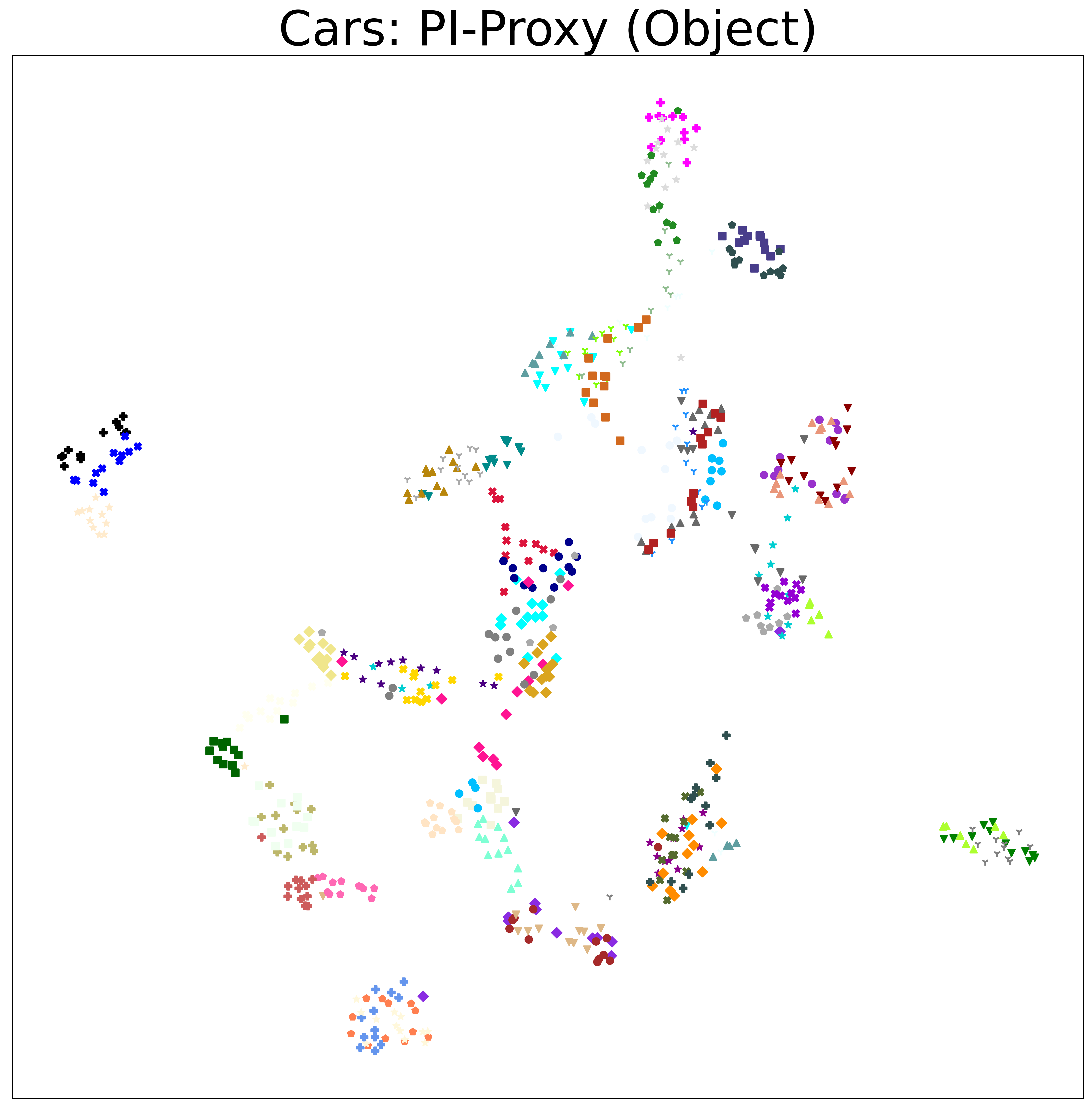}
		\includegraphics[height=0.25\textwidth]{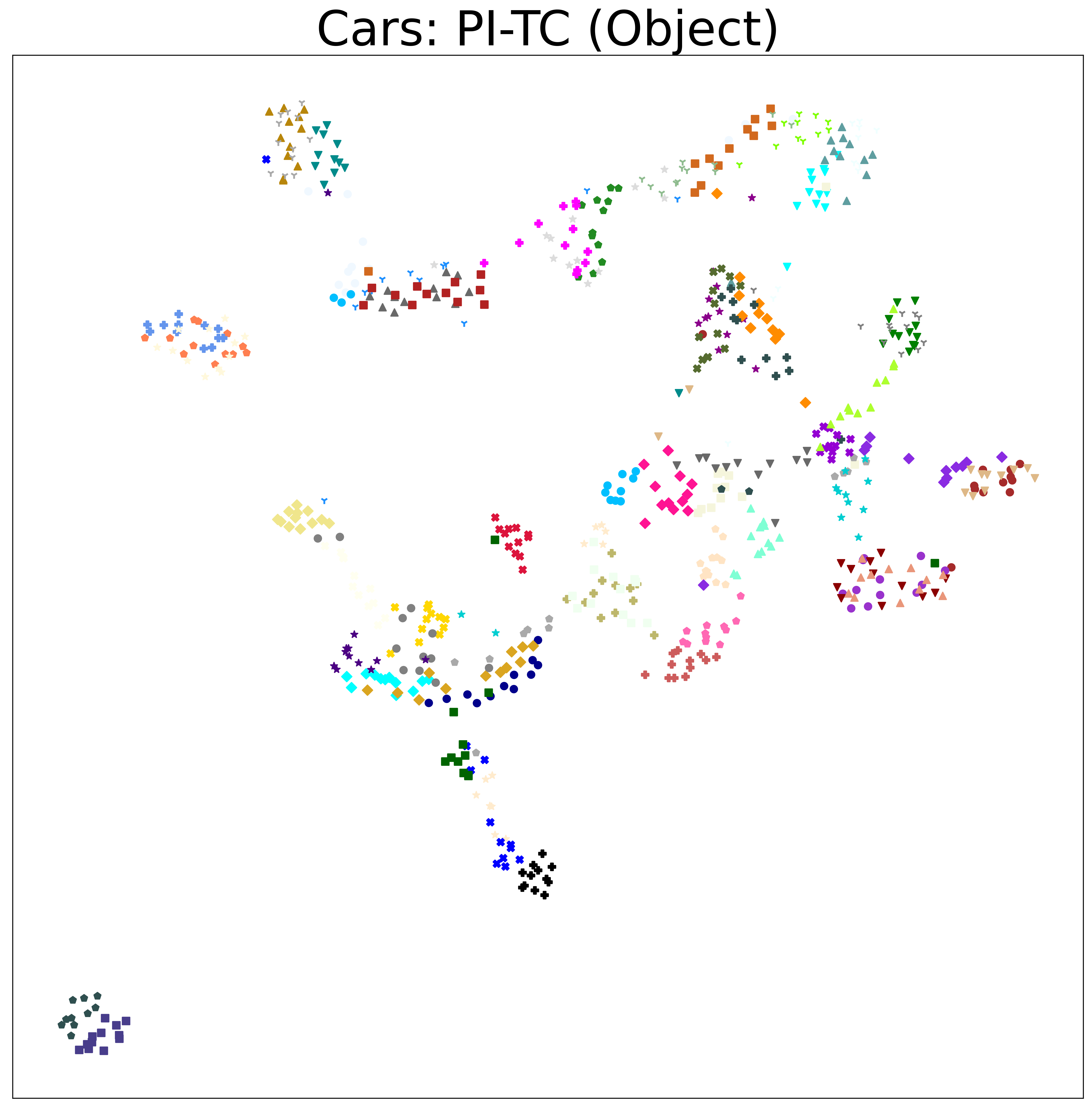}
		\includegraphics[height=0.25\textwidth]{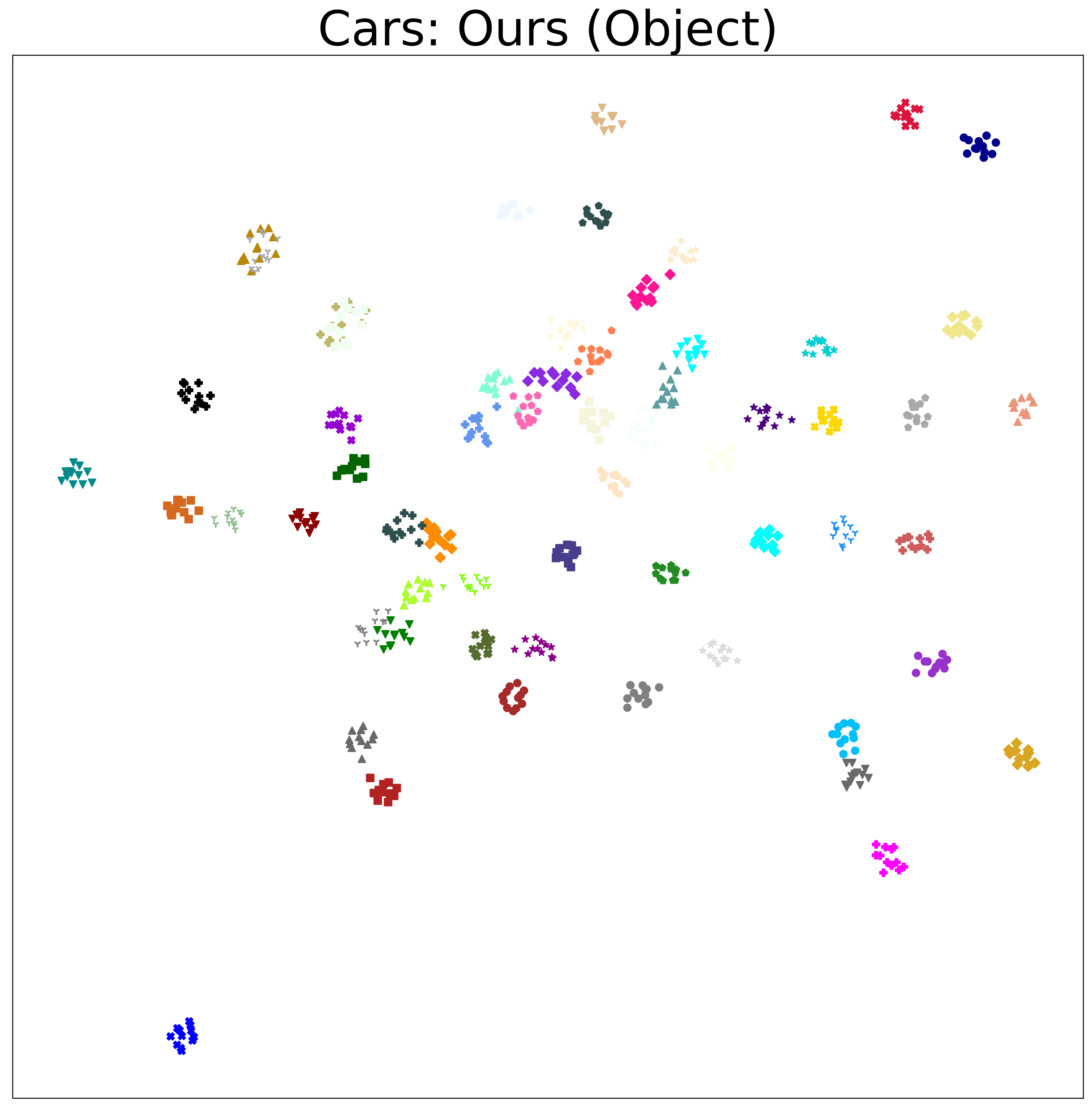}
	%\caption*{\em (b) UMAP visualization}
 \vspace{-0.25in}
 \caption*{\em (b) Object-identity embeddings for 60 car objects (3 objects each from 20 car categories)}
		\includegraphics[height=0.25\textwidth]{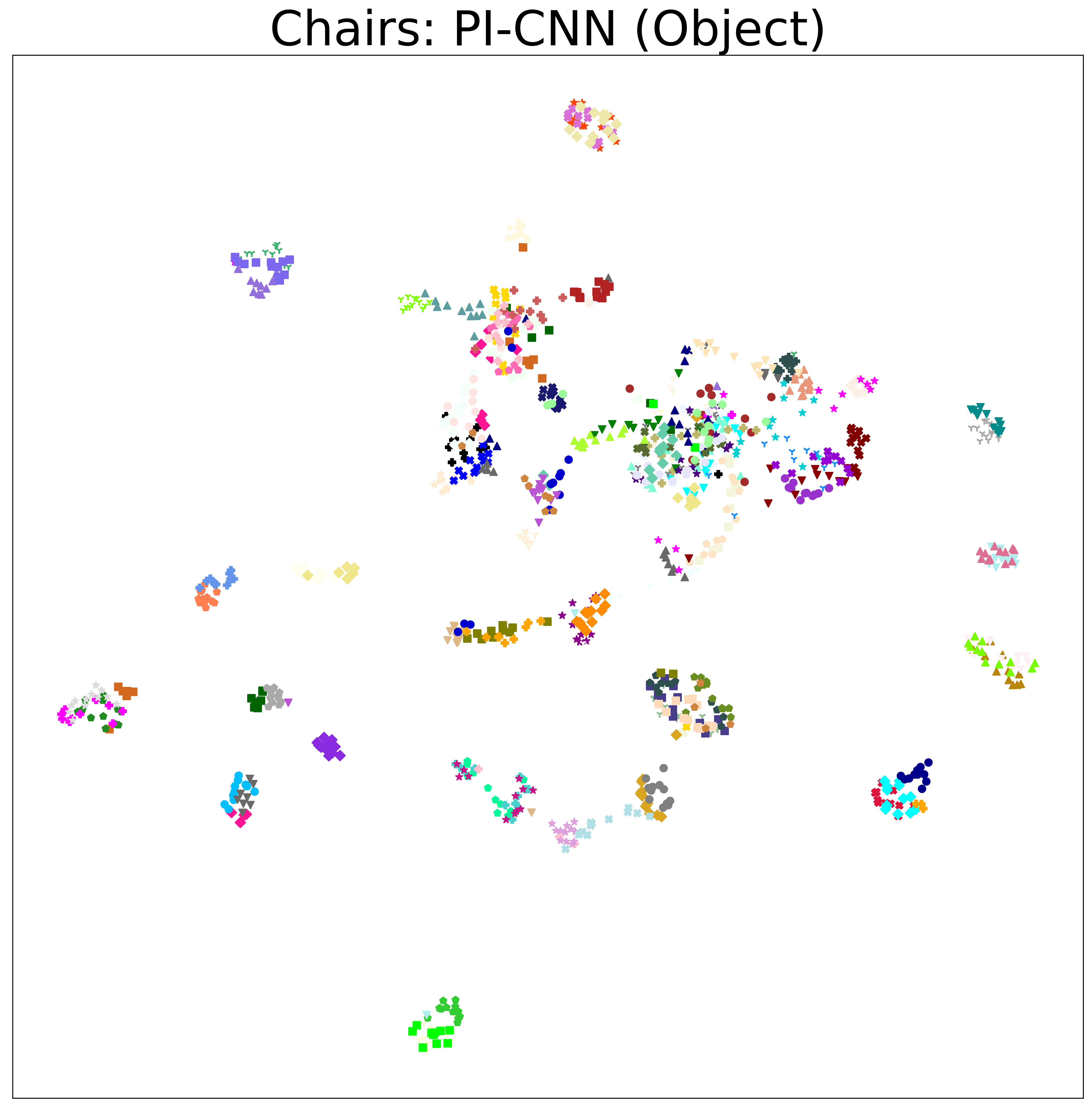}
		\includegraphics[height=0.25\textwidth]{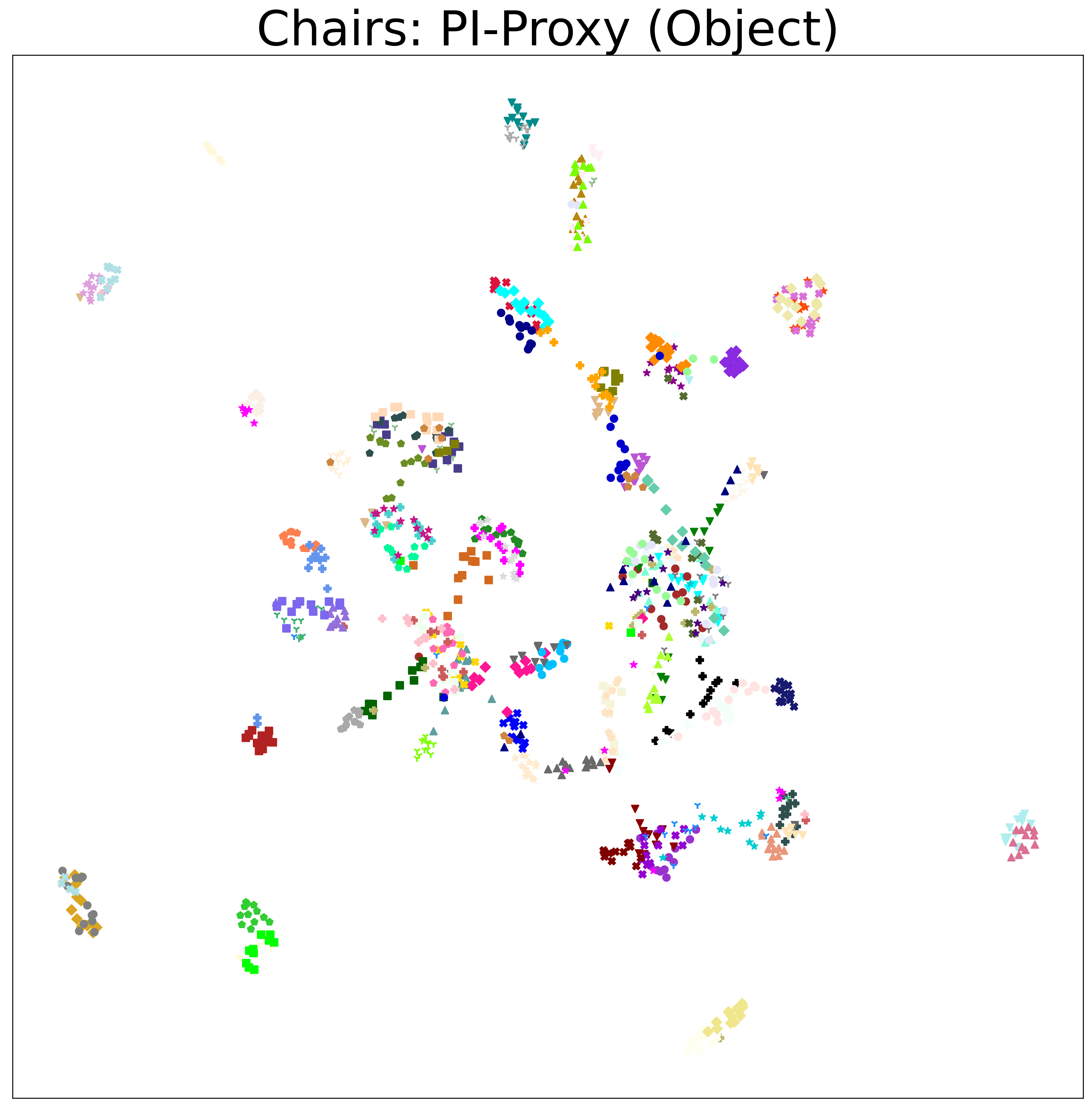}
		\includegraphics[height=0.25\textwidth]{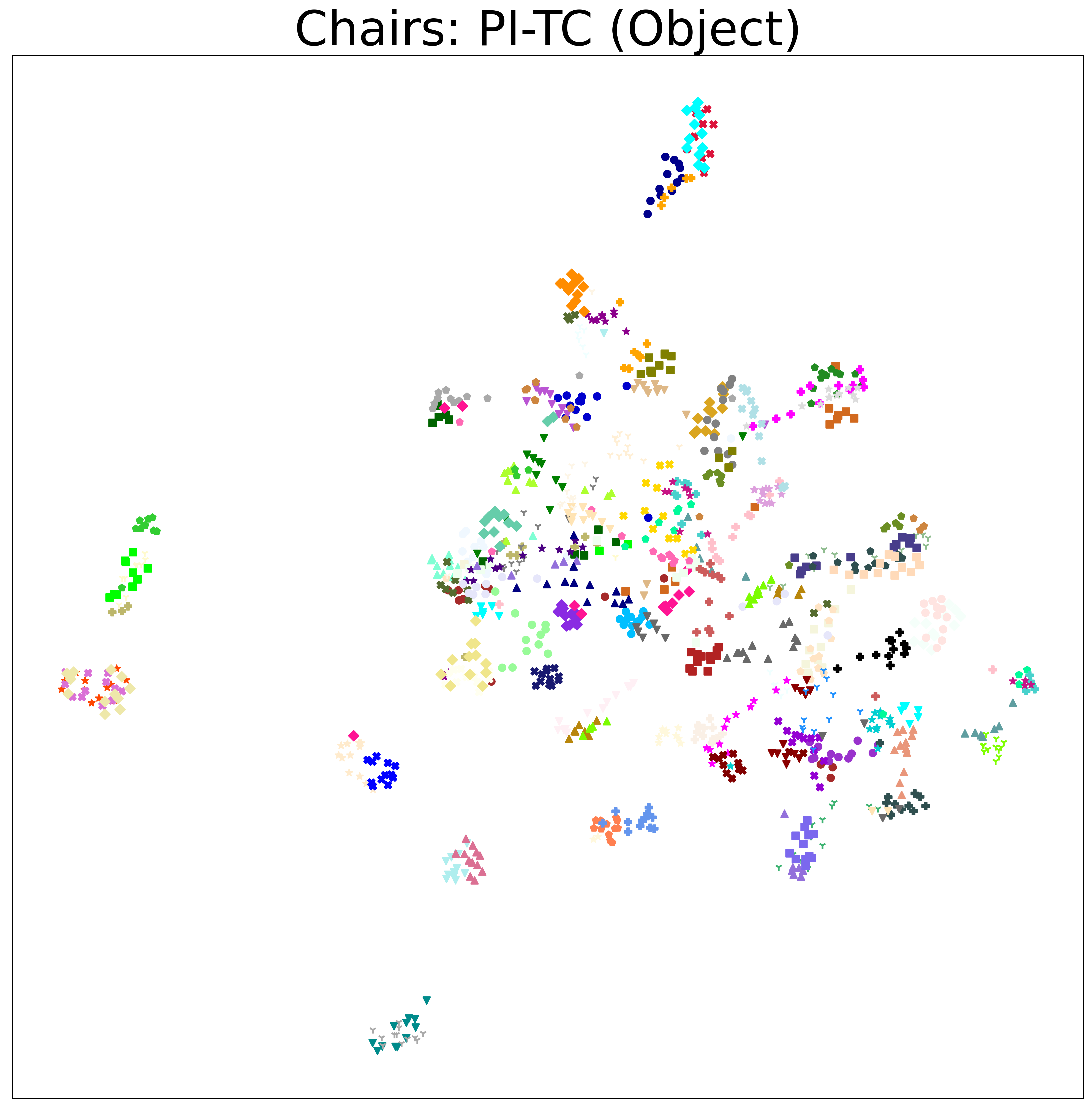}
		\includegraphics[height=0.25\textwidth]{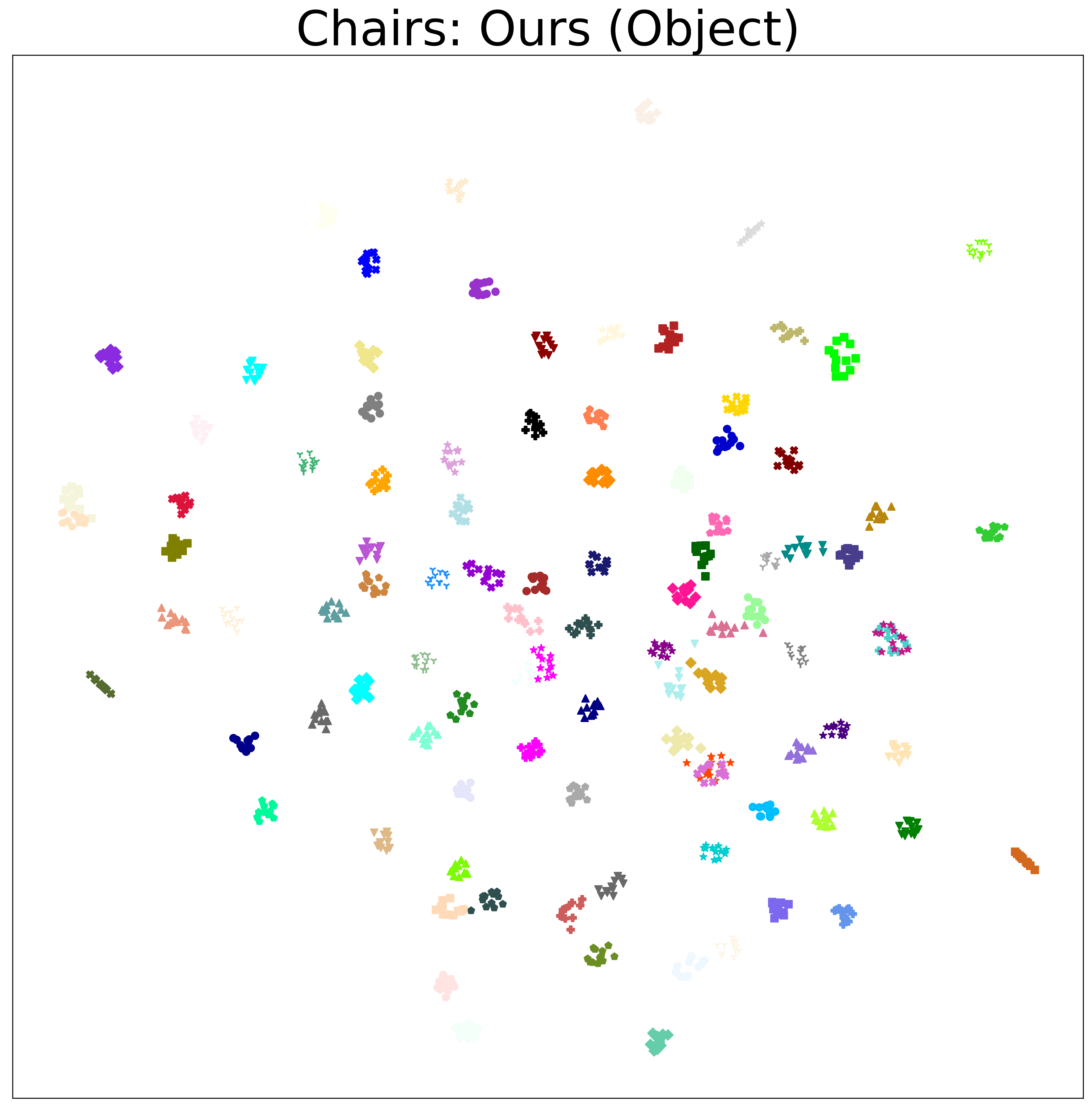}
	%\caption*{\em (b) UMAP visualization}
 \vspace{-0.25in}
 \caption*{\em (c) Object-identity embeddings for 99 chair objects (3 objects each from 33 chair categories)}
 \vspace{-0.1in}
 \caption{\em Comparison of the object embedding space learned for the FG3D test dataset (comprising objects with fine-grained differences from 13 airplane categories in (a), 20 car categories in (b), and 33 chair categories in (c)) by prior pose-invariant methods \cite{PIE2019} and our method (right). Each instance is an object view and each object-identity class is denoted by a unique color and shape. 
It can be observed that our object-identity embeddings are better clustered and separated from other objects as compared to prior methods. 
 }
\label{fig:umap_fg3d}
\end{figure*}

In the supplementary material, we present additional results that could not be reported in detail in the main paper due to space constraints. Our supplemental is organized as follows.
In Sec. \ref{sec:vis_pi}, we present UMAP visualizations of the learned pose-invariant embeddings for the ObjectPI, ModelNet40, and FG3D datasets. In Sec. \ref{sec:abl_piobj}, we present a detailed ablation study of the different components of the proposed pose-invariant object loss. We investigate how the inter-class and intra-class distances for the object-identity classes are optimized in the object embedding space, and further explain how the separation of object-identity classes leads to significant performance improvement on object-level tasks. 
Furthermore, we investigate how the object-identity classes are better separated when learning dual embedding spaces as compared to a single embedding space in Sec. \ref{sec:abl_opt_dist}.
Subsequently in Sec. \ref{sec:emb_dim}, we study the effect of embedding dimensionality on category and object-based classification and retrieval tasks. Next, we illustrate how self-attention captures correlations between different views of an object using multi-view attention maps in Sec. \ref{sec:mv_maps}, and finally present qualitative single-view object retrieval results in Sec. \ref{sec:retr_vis}.

\section{UMAP visualization of pose-invariant embeddings}
\label{sec:vis_pi}

For a qualitative understanding of the effectiveness of our approach, we compare the embeddings generated by the prior pose-invariant methods (specifically, PI-CNN, PI-Proxy, and PI-TC) in \cite{PIE2019} with our method. For this, we use UMAP to project the embeddings into the 2-dimensional space for visualization. In Fig. \ref{fig:umap_mnet40}, we compare the UMAP plots for a subset of the test dataset of ModelNet-40. Since ModelNet-40 has a large number of objects in the test dataset, we choose 100 objects for visualization from five mutually confusing categories such as tables and desks, chairs, stools, and sofas. 
In the category plots, we use five distinct colors to indicate instances from each of the five categories. In the object plots, instances of each object-identity class are indicated by a unique color and shape.

Prior pose-invariant methods (PI-CNN, PI-Proxy, and PI-TC) in \cite{PIE2019} learn a single embedding space. For each of these methods, we show the UMAP visualizations of the same embedding space with the category and object-identity labels in the two subfigures (titled category and object). As mentioned in the paper, prior work focused primarily on learning category-specific embeddings, with the object-to-object variations within each category represented by the variations in the embedding vectors within the same embedding space. 
Specifically, we observe that for PI-CNN and PI-Proxy (in the top row of Fig. \ref{fig:umap_mnet40}), the pose-invariant embeddings for object-identity classes belonging to the same category are not well-separated leading to poor performance on object-based tasks reported in the main paper in Tables \ref{tbl:resultsPICR}, \ref{tbl:fg3d_comp}. PI-TC (bottom-left of Fig. \ref{fig:umap_mnet40}) separates embeddings of the nearest neighbor object-identity classes in the embedding space leading to comparatively better performance. 

In contrast, our method decouples the category and object representations in separate embedding spaces leading to a better separation of both the category and object-identity embeddings, as can be seen in the bottom-right of Fig. \ref{fig:umap_mnet40}.
The most notable difference with prior state-of-the-art is in regards to the learnt object-identity embeddings. Hence, for the other datasets we compare the object-identity embeddings generated by our method and prior pose-invariant methods. In Fig. \ref{fig:umap_objectpi}, we visualize the embeddings for 
the ObjectPI test dataset comprising 98 objects from 25 categories. The FG3D dataset has 66 fine-grained categories that comprise 13 types of airplanes, 20 types of cars, and 33 types of chairs. We sample 3 objects per category and show the object-identity embeddings for the airplane, car, and chair objects separately in Fig. \ref{fig:umap_fg3d} (a), (b), and (c) respectively. For all the datasets, we observe that the object-identity classes are better clustered and separated for our method as compared to prior methods.

We conjecture that our method better separates the object-identity classes for two reasons. First, our method separates confusing instances of objects from the same category that would otherwise be much too close together in the embedding space, as we will explain in detail in Section \ref{sec:abl_piobj}. Second, our method captures category and object-specific discriminative features in separate embedding spaces. Intuitively, this allows us to simultaneously capture common attributes between objects from the same category in the category embedding space and discriminative features to distinguish between them in the object embedding space, as opposed to learning representations to satisfy these conflicting objectives in the same embedding space. This strategy leads to better separability of the object-identity embeddings when learning a dual space as compared to learning a single space, as we will explain in detail in Section \ref{sec:abl_opt_dist}. 

\section{Ablation of Pose-invariant object loss}
\label{sec:abl_piobj}
\begin{figure}[b]
\centering   
\includegraphics[width=0.5\textwidth]{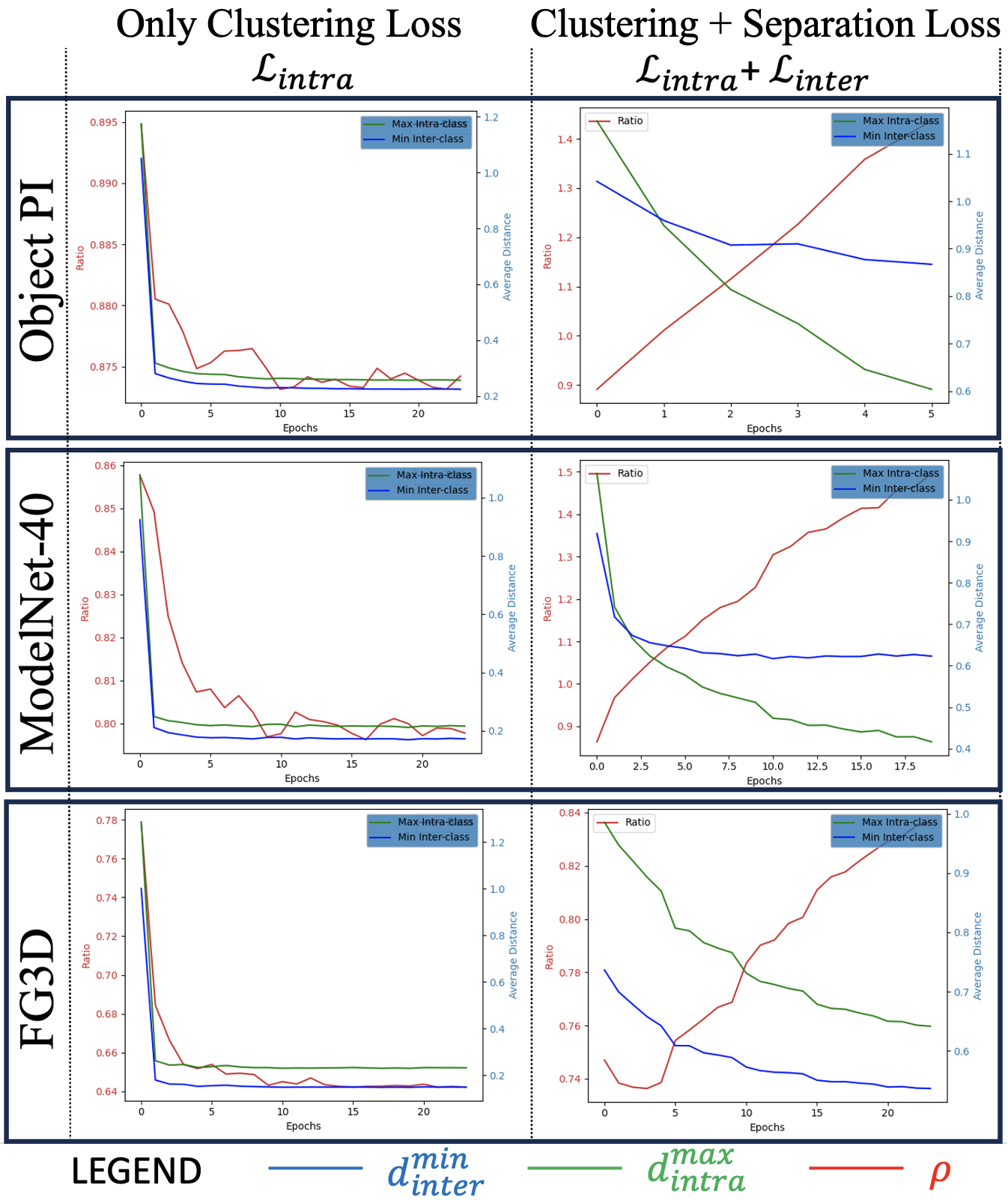}
    \caption{\em This figure illustrates how the minimum inter-class distances between object-identity classes ($d^{min}_{inter}$ in blue), maximum intra-class distances ($d^{max}_{intra}$ in green), and the ratio of the two distances ($\rho$ in red) change as the different components of our pose-invariant object loss are optimized for all three datasets. When training using the clustering loss only, $\rho$ decreases. Whereas, when training using both the losses together, $\rho$ increases indicating better compactness and separability of object-identity classes.}
    \label{fig:lpiobj_abl}
\end{figure}
As explained in Sec. \ref{sec:lit_review}(B) of the main paper, prior approaches primarily focus on clustering the single-view embeddings of each object-identity class close to their multi-view embeddings but do not effectively separate embeddings from different object-identity classes. To ameliorate this, our proposed pose-invariant object loss is designed to separate different object-identity classes, and in this section, we investigate its importance for good performance on object-based tasks. 

Our pose-invariant object loss in Eqn. \ref{eqn:piobj} has two components -- clustering loss ($\mathcal{L}_{intra}$) that reduces the intra-object distances by clustering different views of the same object-identity class, similar to prior approaches. Additionally, we add a separation loss ($\mathcal{L}_{inter}$) that increases the inter-object distances by separating confusing instances of different object-identity classes from the same category. 
To understand the effectiveness of each component, we train the PAN encoder with and without the separation loss. We track how the intra-class and inter-class distances are optimized during training in Fig. \ref{fig:lpiobj_abl}, and also the performance on object recognition and retrieval tasks in Table \ref{tbl:abl_piobj}. 

\begin{table*}[t]
\footnotesize
\centering
\begin{tabular}{@{}|l|l|ccc|cccc|@{}}
\toprule
 &  & \multicolumn{3}{c|}{} & \multicolumn{4}{c|}{Test Performance on Object-level tasks} \\ \cmidrule(l){6-9} 
 &  & \multicolumn{3}{c|}{\multirow{-2}{*}{\begin{tabular}[c]{@{}c@{}}Optimized distances\\ during training\end{tabular}}} & \multicolumn{2}{c|}{Classification (Acc. \%)} & \multicolumn{2}{c|}{Retrieval (mAP \%)} \\ \cmidrule(l){3-9} 
\multirow{-3}{*}{Datasets} & \multirow{-3}{*}{Losses} & $d^{max}_{intra} (\downarrow)$ & $d^{min}_{inter} (\uparrow)$  & $\rho (\uparrow) $  & Single-view & \multicolumn{1}{c|}{Multi-view} & Single-view & Multi-view \\ \midrule
 & $\mathcal{L}_{intra}$ & 0.26 & {\color[HTML]{333333} 0.23} & 0.87 & 84.9 & \multicolumn{1}{c|}{87.8} & 59.8 & 93.2 \\
\multirow{-2}{*}{ObjectPI} & $\mathcal{L}_{intra}+\mathcal{L}_{inter}$ & 0.60 & 0.90 & \textbf{1.50} & \textbf{92.7} & \multicolumn{1}{c|}{\textbf{98.0}} & \textbf{81.0} & \textbf{99.0} \\ \midrule
 & $\mathcal{L}_{intra}$ & 0.22 & {\color[HTML]{333333} 0.18} & 0.80 & 68.5 & \multicolumn{1}{c|}{68.6} & 43.0 & 76.2 \\
\multirow{-2}{*}{ModelNet-40} & $\mathcal{L}_{intra}+\mathcal{L}_{inter}$ & 0.41 & 0.62 & \textbf{1.51} & \textbf{93.7} & \multicolumn{1}{c|}{\textbf{96.9}} & \textbf{84.0} & \textbf{98.2} \\ \midrule
 & $\mathcal{L}_{intra}$ & 0.23 & {\color[HTML]{333333} 0.15} & 0.65 & 20.2 & \multicolumn{1}{c|}{24.4} & 10.4 & 34.7 \\
\multirow{-2}{*}{FG3D} & $\mathcal{L}_{intra}+\mathcal{L}_{inter}$ & 0.63 & 0.53 & \textbf{0.84} & \textbf{83.1} & \multicolumn{1}{c|}{\textbf{91.6}} & \textbf{73.0} & \textbf{95.5} \\ \bottomrule
\end{tabular}
\caption{\em This table shows the maximum intra-class and minimum inter-class distances between object-identity classes after training, and also the test performance on single-view and multi-view object recognition and retrieval tasks for the three datasets, when training with and without the separation loss. }
\label{tbl:abl_piobj}
\end{table*}

\begin{figure}[b]
\centering   
\includegraphics[width=0.5\textwidth]{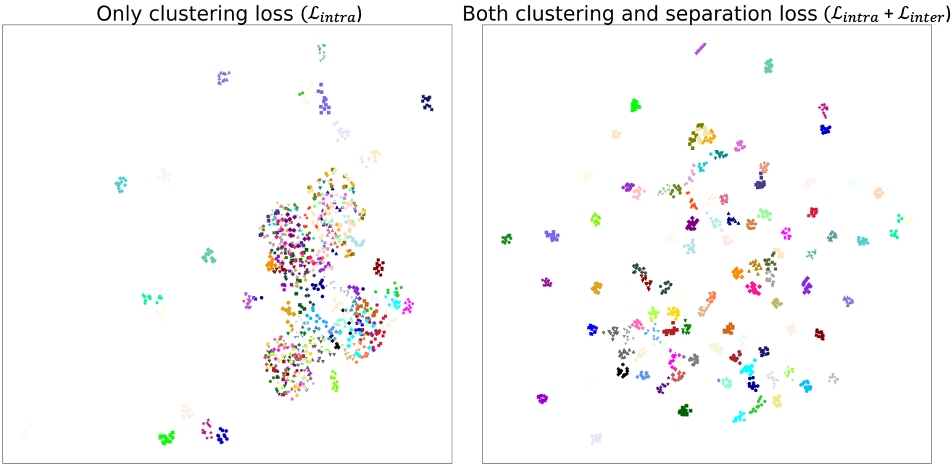}
    \caption{\em We show the object-identity embeddings for a total of 100 objects from 5 categories of ModelNet-40 (20 objects from each category, such as tables, desks, chairs, stools, and sofas). The instances of each object-identity class is indicated by a unique color and shape. This figure illustrates that only clustering embeddings of the same object-identity classes is not sufficient to be able to distinguish between different object-identities, especially when there are many visually similar objects (left). Clustering the different views of the same object-identity classes, and simultaneously separating the different object-identity classes encourages learning more discriminative embeddings (right). }
    \label{fig:sep}
\end{figure}

In Fig. \ref{fig:lpiobj_abl}, we plot the maximum intra-class distance ($d^{max}_{intra}$ in green), and the minimum inter-class distance between object-identity classes from the same category ($d^{min}_{inter}$ in blue) during training to monitor the compactness and separability of object-identity classes respectively. 
These distances are computed using the object-identity embeddings and averaged over all objects. We also plot the ratio $\rho=\frac{d^{min}_{inter}}{d^{max}_{intra}}$ in red. A lower $d^{max}_{intra}$, and higher $d^{min}_{inter}$ and $\rho$ values would indicate embeddings of the same object-identity class are well clustered and separated from embeddings of other object-identity classes from the same category. 

We observe that training using the clustering loss ($\mathcal{L}_{intra}$) reduces the $d^{max}_{intra}$ 
as it encourages clustering different views of the same object-identity together encouraging the network to learn pose-invariant features.
However, only using the clustering loss also reduces $d^{min}_{inter}$, thereby reducing $\rho$ as can be observed in the left of Fig. \ref{fig:lpiobj_abl}.  Therefore, the object-identity classes are not well separated as can be seen in the left of Fig. \ref{fig:sep}, and this results in poor performance on object-level tasks.

 Whereas, when training using the clustering and separation loss jointly ($\mathcal{L}_{intra}+\mathcal{L}_{inter}$), we observe in the right of Fig. \ref{fig:lpiobj_abl} that $\rho$ increases as the $d^{\text{min}}_{\text{inter}}$ decreases at a much slower rate than $d^{\text{max}}_{\text{intra}}$, and $d^{\text{min}}_{\text{inter}}$ eventually converges to a value beyond which it does not decrease substantially, indicating that our loss enforces separability between objects from the same category. 
For all the datasets, adding the separation loss yields significant performance improvement on object-level tasks, as can be seen in Table \ref{tbl:abl_piobj}. This is because it enhances the inter-object separability that allows the encoder to learn more discriminative features to distinguish between visually similar objects. This can be observed in
the right of Fig. \ref{fig:sep}, where each distinct object-identity class (indicated by a unique color and shape) can be more easily distinguished from other object-identity classes.

\section{Optimizing intra-class and inter-class distances in single and dual spaces}
\label{sec:abl_opt_dist}

As mentioned in the previous section, the pose-invariant object loss is designed to simultaneously enhance the inter-class separability and intra-class compactness of object-identity classes. In this section, we study how the distances between the samples in the object embedding space are optimized during training when learning single and dual embedding spaces. 

\begin{figure}[t]
	\centering
        \includegraphics[width=0.5\textwidth]{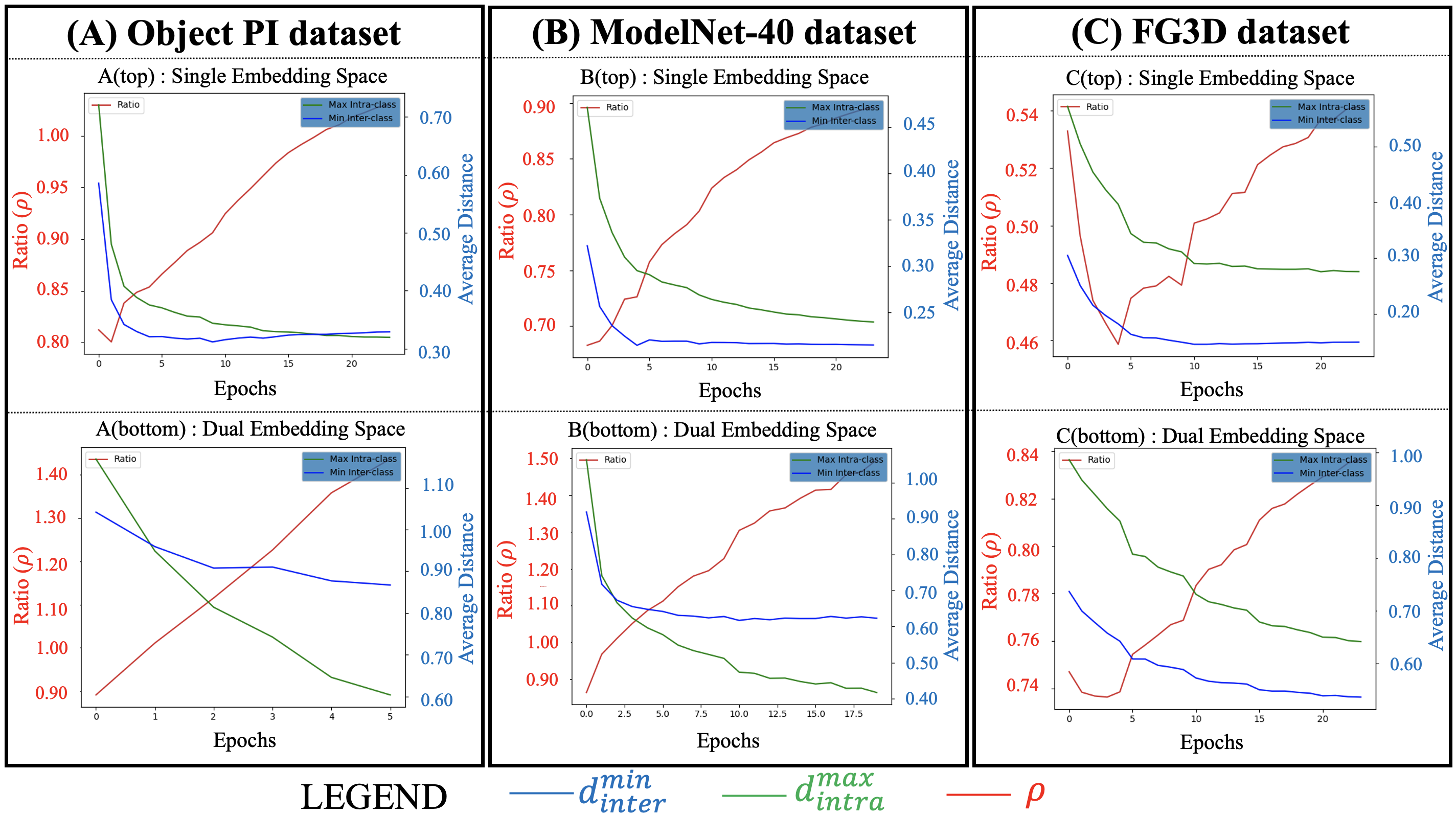}
 \caption{\em For all three datasets, this figure illustrates that our pose-invariant object loss increases the ratio $\rho$ that indicates better separability and compactness of object-identity classes. We observe that the values of the ratio $\rho$ (in red) and the minimum inter-class distance $d^{min}_{inter}$ (in blue) are higher when learning a dual space as compared to a single space. This indicates better separation between object-identity classes when learning in the dual space. }

 \label{fig:ecc_abl_dists}
 \end{figure}

\begin{table*}[b]
\footnotesize
\centering
\begin{tabular}{@{}|l|c|ccc|cccc|@{}}
\toprule
 &  & \multicolumn{3}{c|}{} & \multicolumn{4}{c|}{Test Performance on Object-level tasks} \\ \cmidrule(l){6-9} 
 &  & \multicolumn{3}{c|}{\multirow{-2}{*}{\begin{tabular}[c]{@{}c@{}}Optimized distances\\ during training\end{tabular}}} & \multicolumn{2}{c|}{Classification (Acc. \%)} & \multicolumn{2}{c|}{Retrieval (mAP \%)} \\ \cmidrule(l){3-9} 
\multirow{-3}{*}{Datasets} & \multirow{-3}{*}{\begin{tabular}[c]{@{}c@{}}Embedding \\ Space\end{tabular}} & $d^{max}_{intra} (\downarrow)$ & $d^{min}_{inter} (\uparrow)$  & $\rho (\uparrow) $ & Single-view & \multicolumn{1}{c|}{Multi-view} & Single-view & Multi-view \\ \midrule
 & Single & {\color[HTML]{333333} 0.32} & {\color[HTML]{333333} 0.33} & {\color[HTML]{333333} 1.03} & {\color[HTML]{333333} 88.5} & \multicolumn{1}{c|}{{\color[HTML]{333333} \textbf{98.0}}} & {\color[HTML]{333333} 68.5} & {\color[HTML]{333333} 98.9} \\
\multirow{-2}{*}{ObjectPI} & Dual & {\color[HTML]{333333} 0.60} & {\color[HTML]{333333} \textbf{0.90}} & {\color[HTML]{333333} \textbf{1.50}} & {\color[HTML]{333333} \textbf{92.7}} & \multicolumn{1}{c|}{{\color[HTML]{333333} \textbf{98.0}}} & {\color[HTML]{333333} \textbf{81.0}} & {\color[HTML]{333333} \textbf{99.0}} \\ \midrule
 & Single & {\color[HTML]{333333} 0.24} & {\color[HTML]{333333} 0.22} & {\color[HTML]{333333} 0.92} & {\color[HTML]{333333} 81.2} & \multicolumn{1}{c|}{{\color[HTML]{333333} 85.6}} & {\color[HTML]{333333} 59.2} & {\color[HTML]{333333} 90.4} \\
\multirow{-2}{*}{ModelNet-40} & Dual & {\color[HTML]{333333} 0.41} & {\color[HTML]{333333} \textbf{0.62}} & {\color[HTML]{333333} \textbf{1.51}} & {\color[HTML]{333333} \textbf{93.7}} & \multicolumn{1}{c|}{{\color[HTML]{333333} \textbf{96.9}}} & {\color[HTML]{333333} \textbf{84.0}} & {\color[HTML]{333333} \textbf{98.2}} \\ \midrule
 & Single & {\color[HTML]{333333} 0.29} & {\color[HTML]{333333} 0.16} & {\color[HTML]{333333} 0.55} & {\color[HTML]{333333} 26.2} & \multicolumn{1}{c|}{{\color[HTML]{333333} 31.0}} & {\color[HTML]{333333} 15.7} & {\color[HTML]{333333} 42.9} \\
\multirow{-2}{*}{FG3D} & Dual & {\color[HTML]{333333} 0.63} & {\color[HTML]{333333} \textbf{0.53}} & {\color[HTML]{333333} \textbf{0.84}} & {\color[HTML]{333333} \textbf{83.1}} & \multicolumn{1}{c|}{{\color[HTML]{333333} \textbf{91.6}}} & {\color[HTML]{333333} \textbf{73.0}} & {\color[HTML]{333333} \textbf{95.5}} \\ \bottomrule
\end{tabular}
\caption{\em This table shows the maximum intra-class and minimum inter-class distances between object-identity classes after training when learning a single and dual embedding space. We observe that the $d^{min}_{inter}$ and $\rho$ values are higher in the dual embedding space indicating better separability of object-identity classes in the object embedding space. This yields better performance on object-level tasks for all the three datasets. }
\label{tbl:abl_single_dual}
\end{table*}

For comparison, we show how these distances are optimized when learning representations in a single and dual embedding spaces, at the top and bottom of Fig. \ref{fig:ecc_abl_dists}(A), (B), and (C) for the ObjectPI, ModelNet-40, and FG3D datasets respectively. We observe that the values of $\rho$ and the minimum inter-object distances ($d^{\text{min}}_{\text{inter}}$) are much higher in the dual space, which indicates that the object embeddings in the dual embedding space are better separated than those in the single embedding space, leading to better performance on object-based tasks, as shown in Table \ref{tbl:abl_single_dual}.

\begin{figure}[t]
	\centering
        \includegraphics[width=0.5\textwidth]{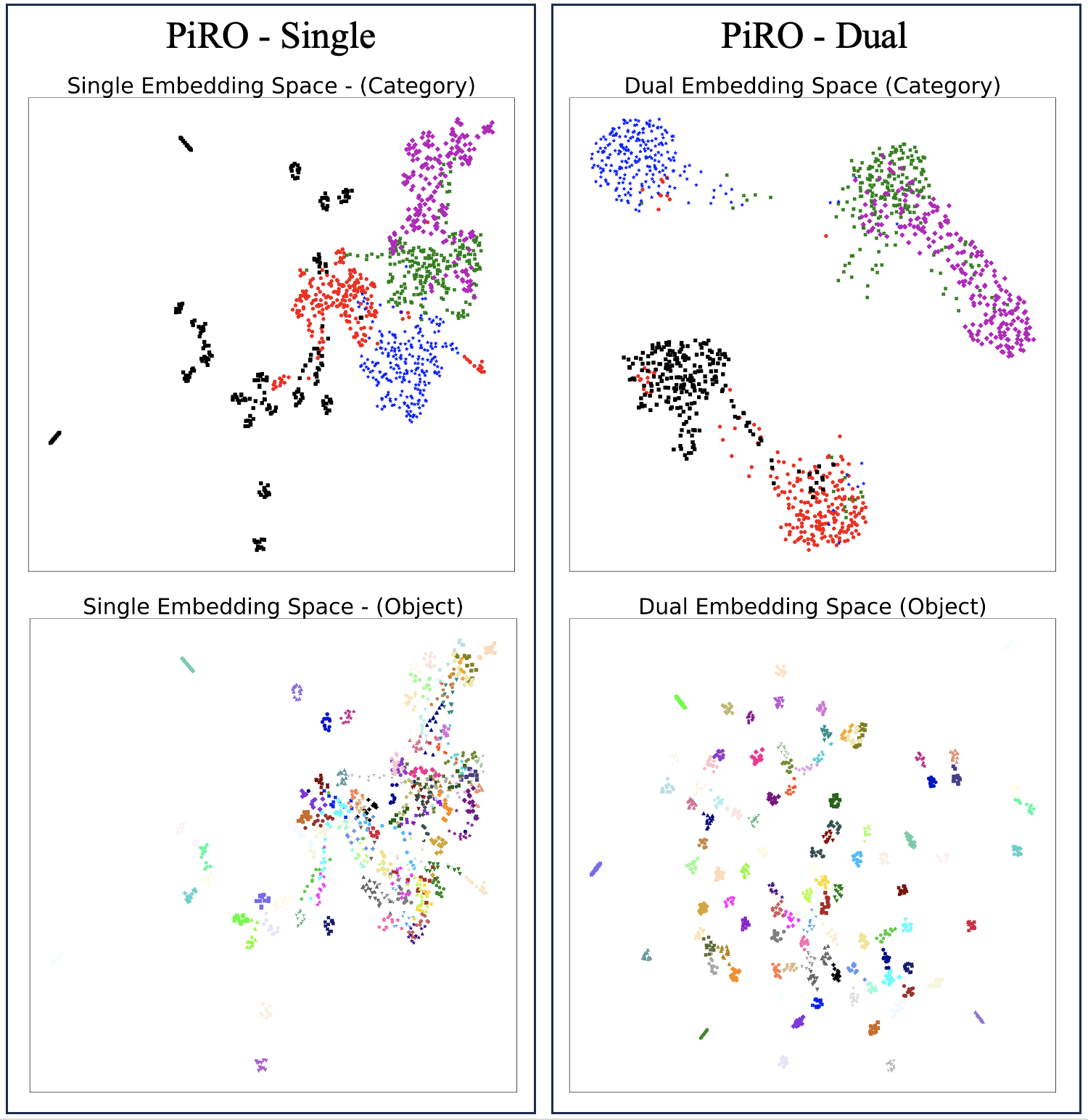}
        \vspace{-0.1in}
 \caption{\em UMAP visualization of the 100 objects from 5 different categories of ModelNet-40 for the single embedding space (left) and dual embedding space (right). The figure illustrates that decoupling the category and object-identity representations in separate spaces leads to better separability between categories in the category embedding space and object-identity classes in the object embedding space (right) as compared to learning representations in the same embedding space (left). }
 \label{fig:umap_single_dual}
 \end{figure}

Fig. \ref{fig:umap_single_dual} illustrates this effect using UMAP visualizations of the embeddings in single and dual spaces. As mentioned in the paper, we jointly train our encoder using pose-invariant category and object-based losses. In the single embedding space, category-based losses aim to cluster embeddings of object-identity classes from the same category together, and in the same embedding space, the object-based loss aims to separate different object-identity classes from the same category. Due to these conflicting objectives in the same embedding space, object-identity classes are not separated well, as can be seen in Fig. \ref{fig:umap_single_dual} (left). In the dual space, the category and object representations are decoupled, and the category and object losses optimize the distances in the separate embedding spaces. As can be seen in Fig. \ref{fig:umap_single_dual} (right), the object and category embeddings are much better separated and we learn more discriminative embeddings overall in the dual space. This leads to significant performance improvements on object-based tasks, as can be seen in Table \ref{tbl:abl_single_dual}. 
\section{Embedding dimensionality}
\label{sec:emb_dim}
\begin{figure}
	\centering
		\centering
		\includegraphics[width=0.5\textwidth]{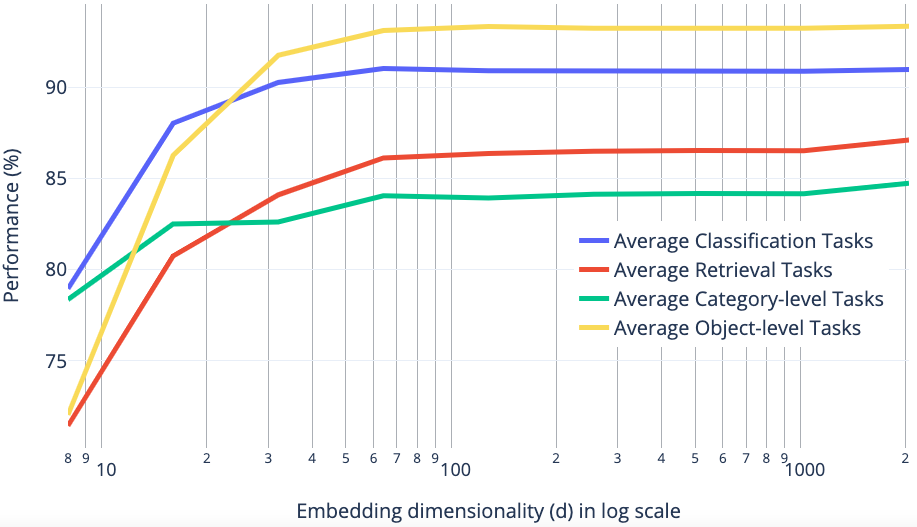}
        \caption*{(a) ModelNet-40 dataset}
		\includegraphics[width=0.5\textwidth]{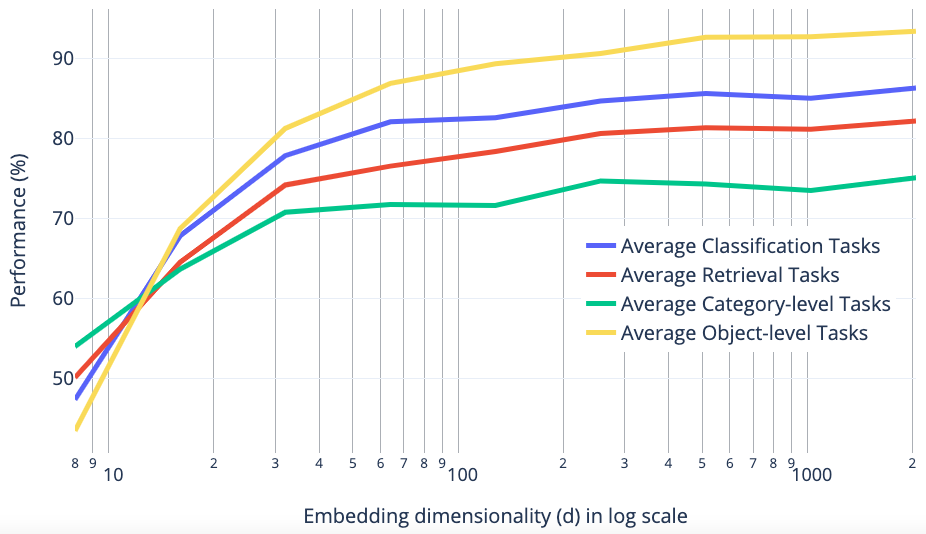}
        \caption*{(b) ObjectPI dataset}
	\caption{\em This graph illustrates the effect of embedding dimensionality on Average Classification and Retrieval performance for category and object-based tasks for ModelNet-40 (top) and ObjectPI (bottom) datasets. 
 }
\label{fig:embed_dim}
\end{figure}
For this experiment, we varied the embedding dimensionality from $d=8, 16, \cdots, 2048$ for the category and object embedding space. We measured the performance of our method in terms of the average classification and retrieval performance as well as average performance on category-based and object-based tasks. 

From Fig. \ref{fig:embed_dim}, we observe that the performance on all four metrics improves with an increase in embedding dimensionality but beyond a certain embedding dimension, the performance only improves marginally. For the ModelNet-40 dataset, we observe that $d=64$ for category-based tasks and $d=128$ for object-based tasks is sufficient. For the ObjectPI dataset, we observe that $d=256$ and $d=512$ are sufficient for category-based and object-based tasks. We conjecture that higher embedding dimensionality is required for ObjectPI than ModelNet-40 as the embeddings for ObjectPI need to additionally capture color and texture information instead of just shape information for ModelNet-40. 
In general, the embedding dimensionality required for good performance on object-based tasks is higher than on category-based tasks as object embeddings need to capture more fine-grained details to differentiate between objects. 

\begin{figure}[t]
    \centering
    \includegraphics[width=0.5\textwidth]{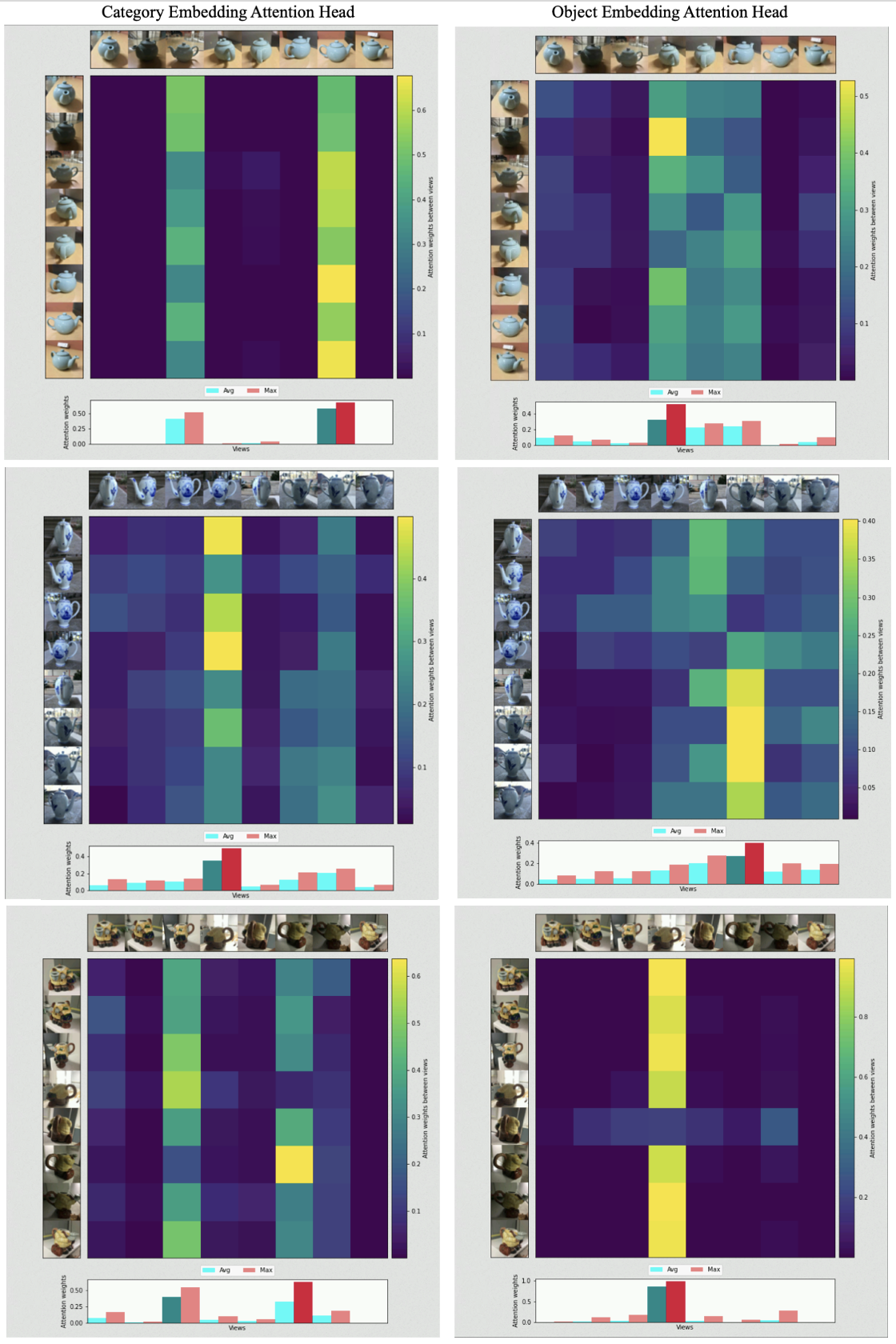}
    \caption{\em Visualization of multi-view attention maps for the category and object self-attention layers. }
    \label{fig:attention heads}
\end{figure}
\section {Multi-view attention maps}
\label{sec:mv_maps}
We plot the attention weights for the self-attention layers for the object and category embeddings in Figure \ref{fig:attention heads}.  We observe that for category embeddings, the attention weights are higher for representative views that capture the overall shape of the kettle. All views of the object are correlated to these representative views. 
For object embeddings, we observe that the attention weights are higher for the views that capture attributes related to the handle. This is possibly because the different kettles in the dataset have variations in the location (from the top or side) and shape of the handle. 

\section{Qualitative Retrieval Results}
\label{sec:retr_vis}
\begin{figure*}[b]
    \centering
    \includegraphics[width=0.9\textwidth]{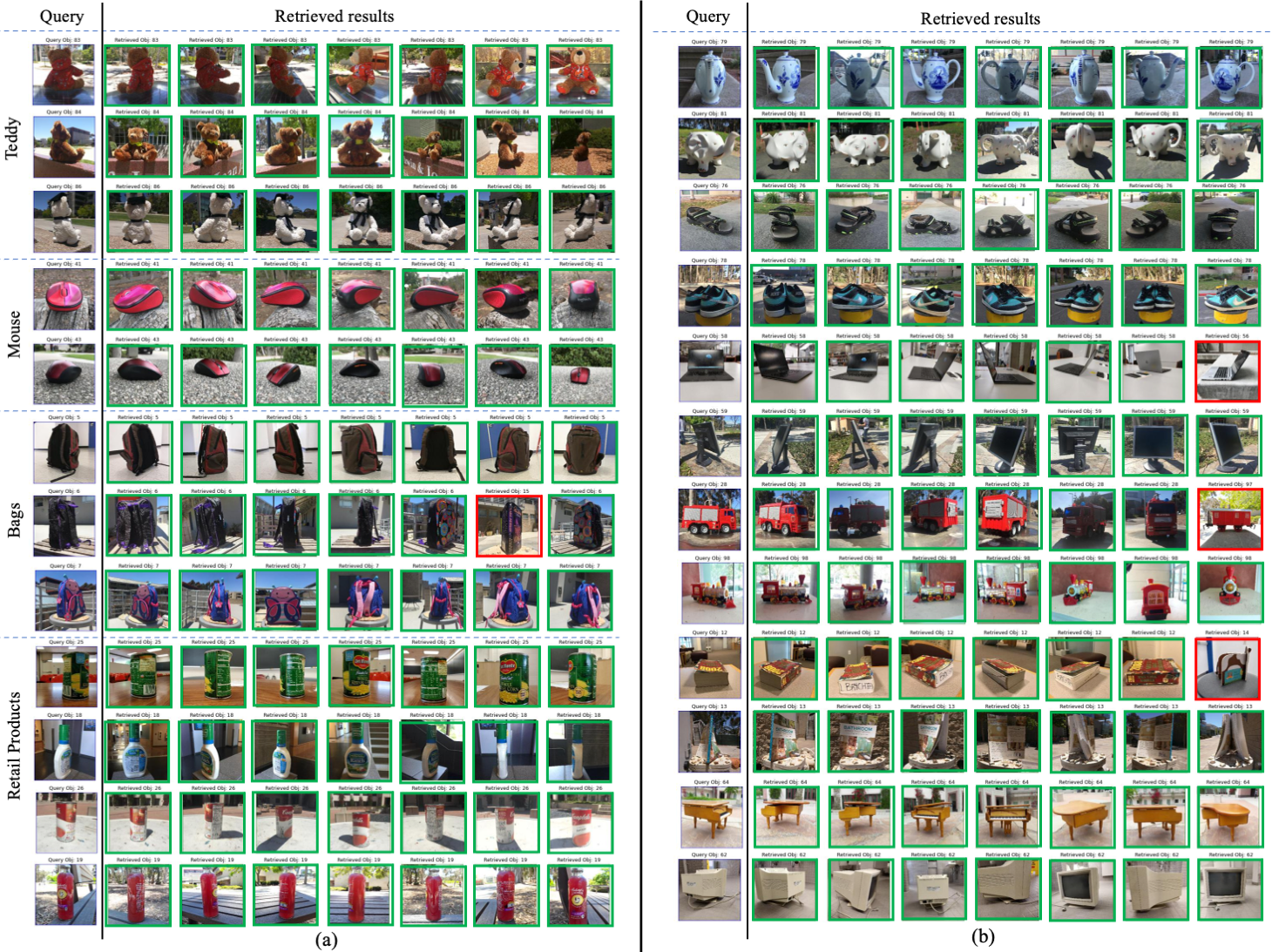}
    \caption{\em This figure shows our object retrieval results for the Object PI dataset. Given a single view query from an arbitrary pose on the left, the top-7 retrieved results are shown on the right in each row. Green bounding boxes indicate correct retrieval results and red boxes indicate incorrect results. In (a), we demonstrate that our framework can identify retrieve other views of the same object despite having similar objects in the test dataset. In (b), we demonstrate, despite significant appearance changes under various pose transformations for different everyday objects, our framework can retrieve objects accurately.  }
    \label{fig:svor_oowl}
\end{figure*}

\begin{figure}[h!]
    \centering
        \includegraphics[width=0.45\textwidth]{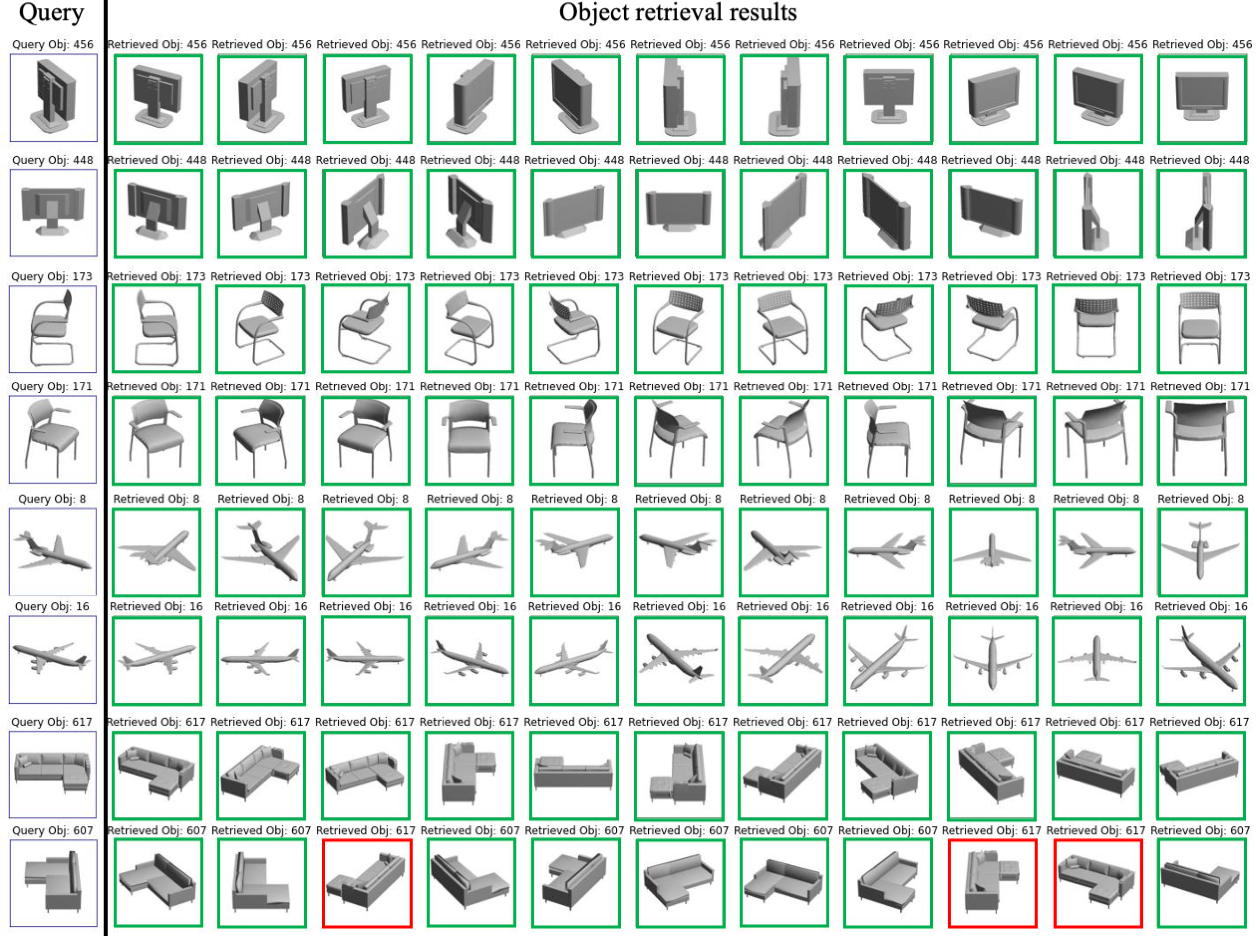}
    \caption{\em For a single-view query in each row, the retrieved images of other views of the same object are shown on the right for the ModelNet40 dataset. The green and red bounding boxes indicate correct and wrong results respectively.}
    \label{fig:svor_mnet}
\end{figure}

We show some qualitative object retrieval results on ObjectPI and ModelNet40 datasets in Figs. \ref{fig:svor_oowl} and \ref{fig:svor_mnet} respectively. 
The single-view object retrieval results show that given an 
arbitrary view of the object, our method can retrieve the other views of the same object correctly in Figs. \ref{fig:svor_oowl} and \ref{fig:svor_mnet}. Despite variability in object appearance from different viewpoints, the presence of similar objects in the database as well as deformable objects (such as books, clothing, and so on), our method can retrieve objects with high precision.

\end{document}